\documentclass{bmvc2k}

\title{Part-based Graph Convolutional Network for Action Recognition}

\addauthor{Kalpit Thakkar}{kalpit.thakkar@research.iiit.ac.in}{1}
\addauthor{P J Narayanan}{pjn@iiit.ac.in}{1}

\addinstitution{
Center for Visual Information Technology (CVIT),
Kohli Center for Intelligent Systems (KCIS),\\
IIIT Hyderabad,
India
}

\runninghead{Kalpit Thakkar, P J Narayanan}{Part-based GCN For Action Recognition}

\def\etal{\emph{et al}\bmvaOneDot}

\usepackage[compact]{titlesec} 

\begin{document}

\maketitle

\begin{abstract}
Human actions comprise of joint motion of articulated body parts or ``gestures''. Human skeleton is intuitively represented as a sparse graph with joints as nodes and natural connections between them as edges. Graph convolutional networks have been used to recognize actions from skeletal videos. We introduce a part-based graph convolutional network (PB-GCN) for this task, inspired by Deformable Part-based Models (DPMs). We divide the skeleton graph into four subgraphs with joints shared across them and learn a recognition model using a part-based graph convolutional network. We show that such a model improves performance of recognition, compared to a model using entire skeleton graph. Instead of using 3D joint coordinates as node features, we show that using relative coordinates and temporal displacements boosts performance. Our model achieves state-of-the-art performance on two challenging benchmark datasets NTURGB+D and HDM05, for skeletal action recognition.
\end{abstract}

\section{Introduction}
\label{sec:intro}
Recognizing human actions in videos is necessary for understanding them. Video modalities such as RGB, depth and skeleton provide different types of information for understanding human actions. The S-video (or Skeletal
modality) provides 3D joint locations, which is a relatively high level information compared to RGB or depth. With the release of several multi-modal datasets \cite{Shahroudy_2016_CVPR,liu2017pku,7350781}, action recognition from S-video has gained significant traction recently \cite{liu2016spatio,song2017end,liu2017global,zhang2017geometric,ke2017new}.

Graph convolutions \cite{niepert2016learning,defferrard2016convolutional,kipf2016semi} have been used to learn high level features from arbitrary graph structure. State-of-the-art action recognition from S-videos \cite{yan2018spatial,li2018spatio} use graph convolutions, wherein the whole skeleton is treated as a single graph. It is, however, natural to think of human skeleton as a combination of multiple body parts. A body-part based representation can learn the importance of each part and their relations across space and time. We present a model using part-based graph convolutional network for recognizing actions from S-videos, using a novel part-based graph convolution scheme. The model attains better performance for recognition than a model entire skeleton as a single graph. Current models for skeletal action recognition \cite{yan2018spatial,li2018spatio} use 3D coordinates as features at each vertex. Geometric features such as relative joint coordinates and motion features such as temporal displacements can be more informative for action recognition. Optical flow helps in action recognition from RGB videos \cite{wang2016temporal} and Manhattan line map helps in generating 3D layout from single image \cite{zou2018layoutnet}. Geometric feature \cite{zhang2017geometric} and kinematic features \cite{zanfir2013moving} have been used for skeletal action recognition before. Inspired by these observations, we use a geometric feature that encodes relative joint coordinates and motion feature that encodes temporal displacements at each vertex in our part-based graph convolution model to significant impact.

\begin{figure}[t]
\begin{center}
    \begin{tabular}{cc}
        \begin{tabular}{cc}
            \includegraphics[width=0.15\textwidth,height=0.25\textwidth]{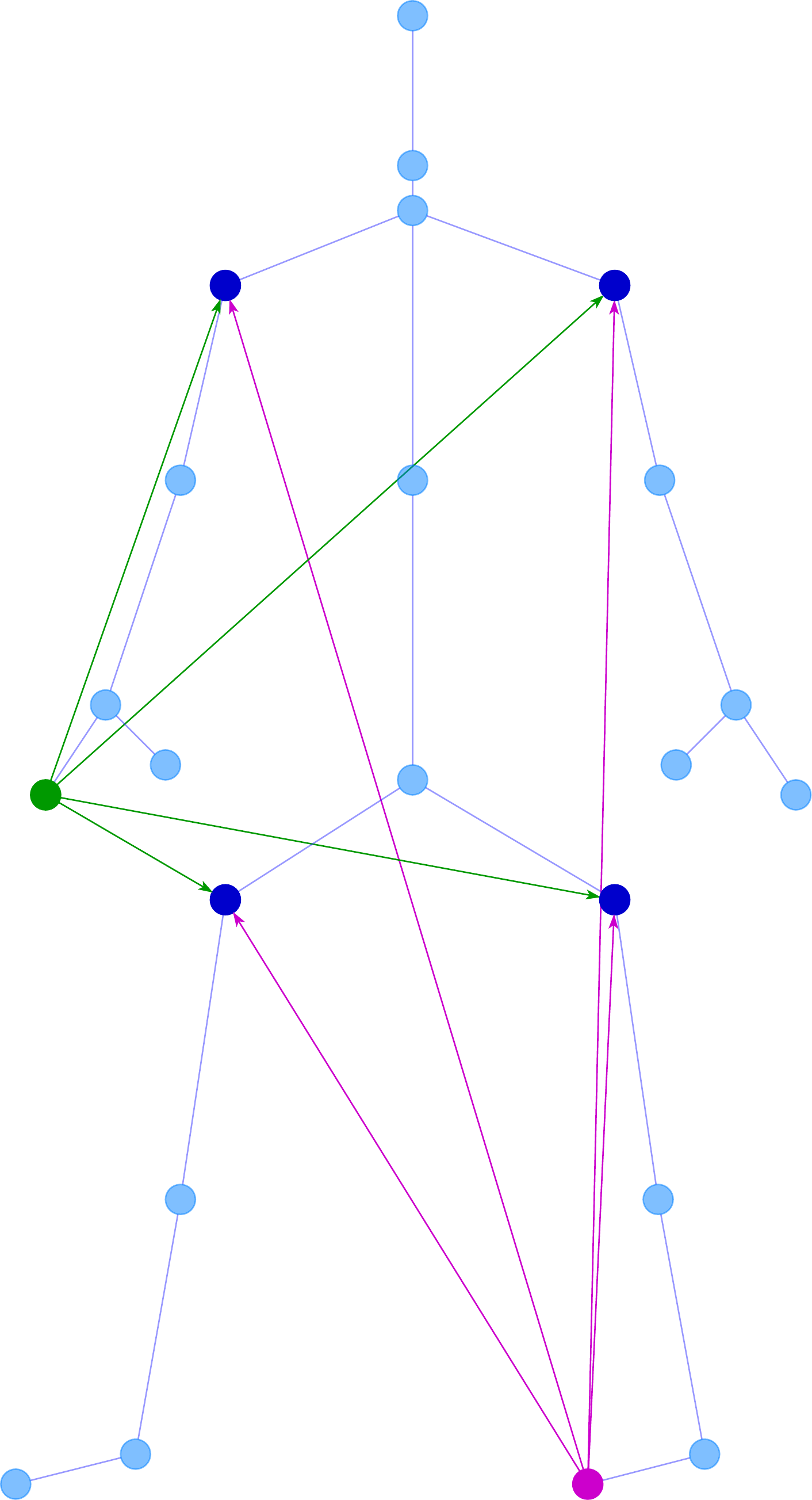} &
            \includegraphics[width=0.15\textwidth,height=0.25\textwidth]{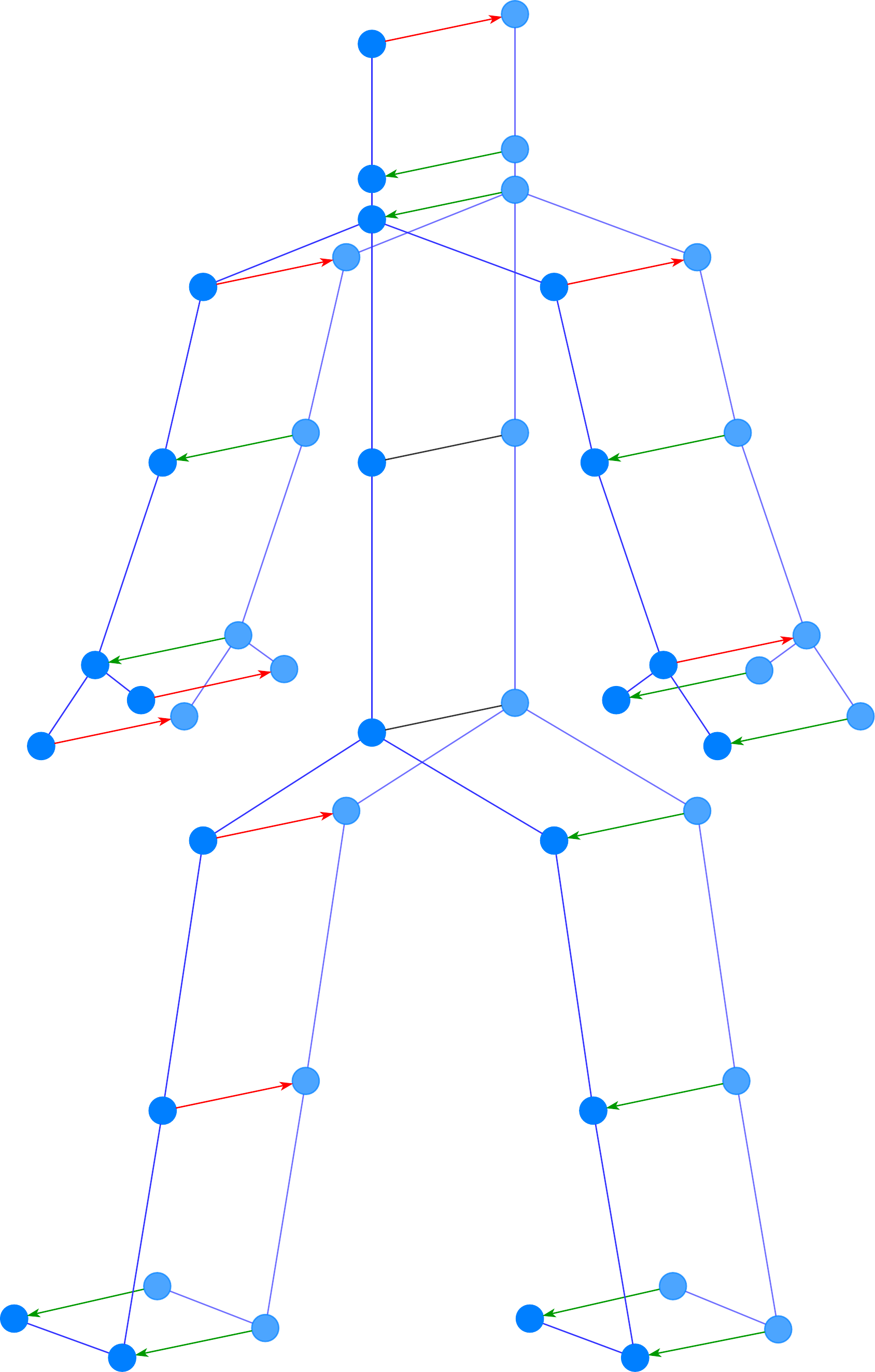} \\
            \footnotesize{Relative coordinates} & \footnotesize{Temporal displacements} \\
        \end{tabular}
        &
        \begin{tabular}{c}
            \includegraphics[width=0.15\textwidth,height=0.25\textwidth]{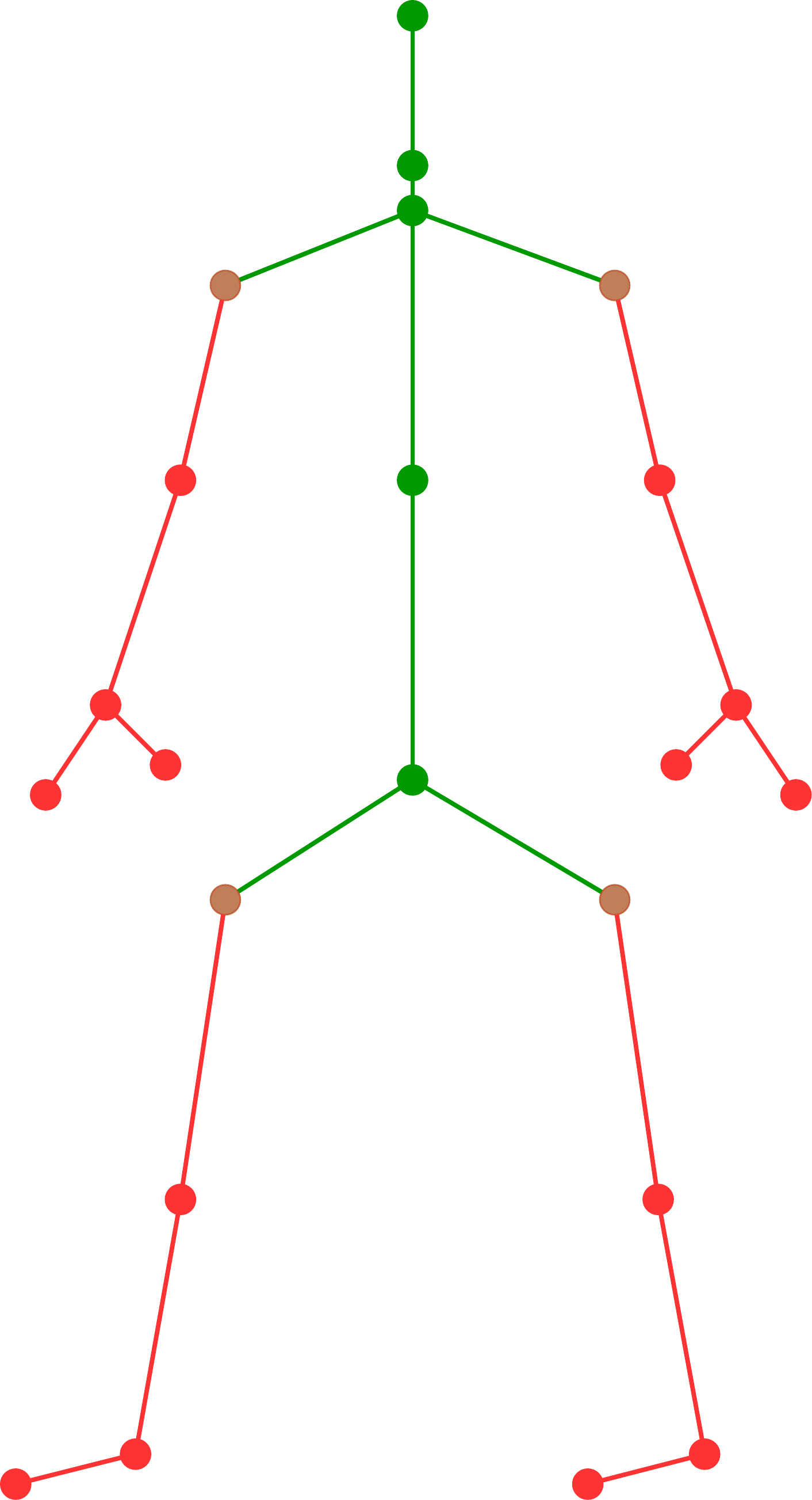} \\
            \footnotesize{Axial (Green) \& Appendicular (Red) skeletons}
        \end{tabular} \\
        \small{(a) Geometric \& Kinematic Features} & \small{(b) Two parts} \\
        \begin{tabular}{cc}
            \includegraphics[width=0.10\textwidth]{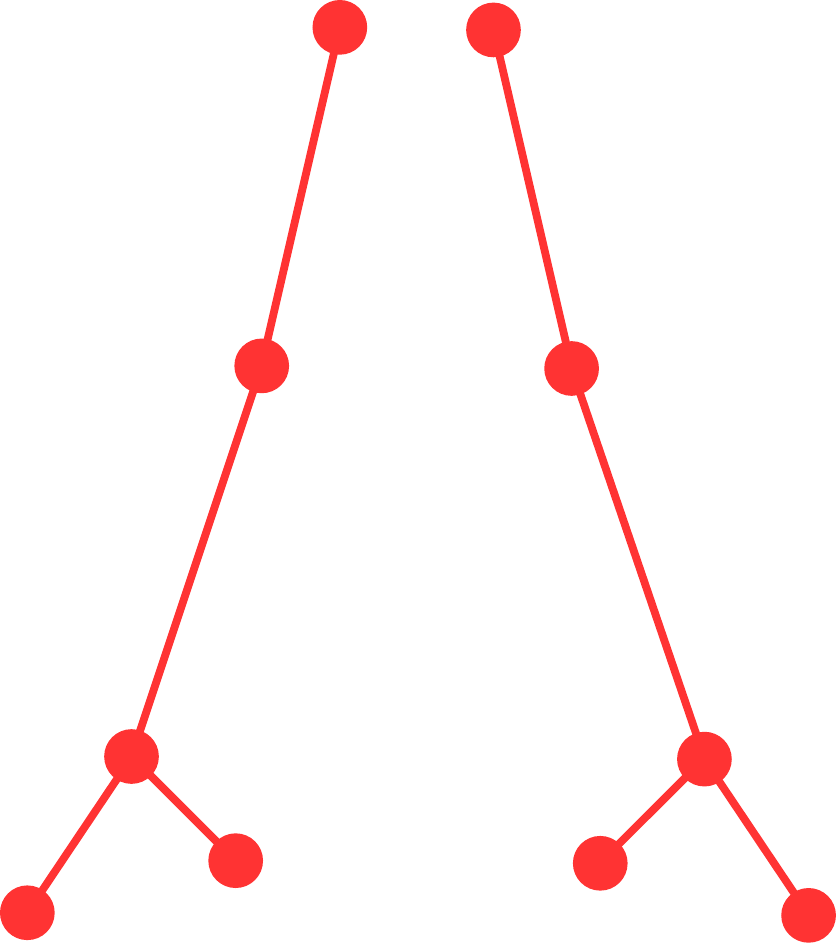} &
            \includegraphics[width=0.10\textwidth]{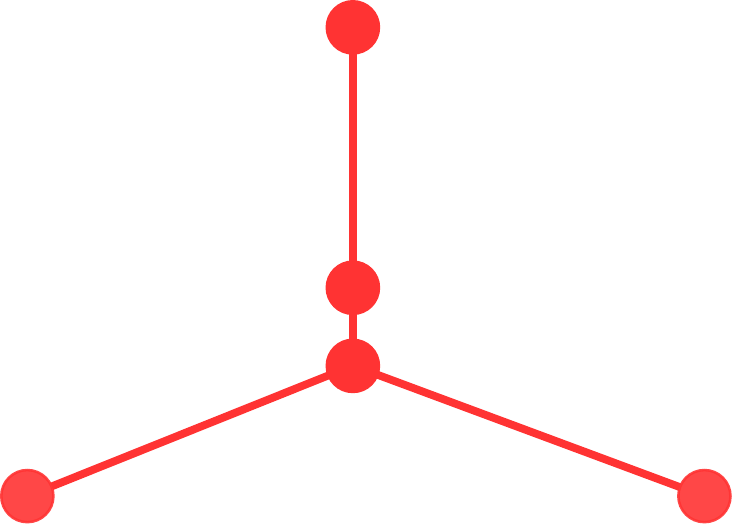} \\
            \footnotesize{Upper} & \footnotesize{Upper} \\
            \footnotesize{Appendicular} & \footnotesize{Axial} \\
            \includegraphics[width=0.10\textwidth]{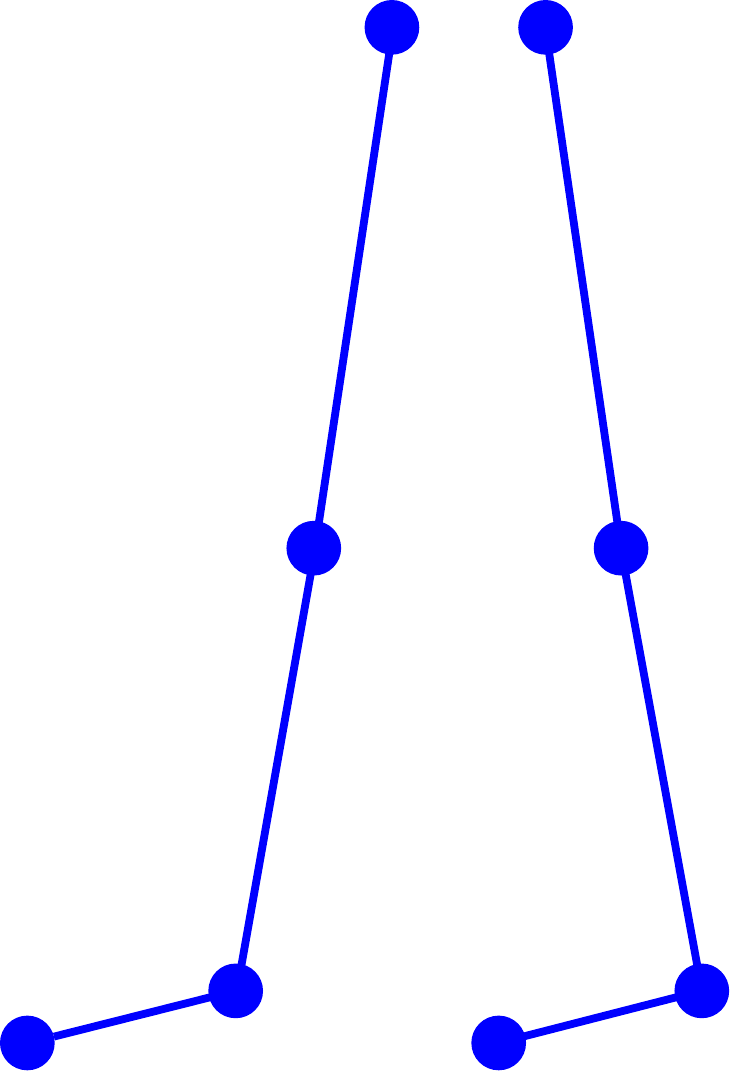} &
            \includegraphics[width=0.10\textwidth]{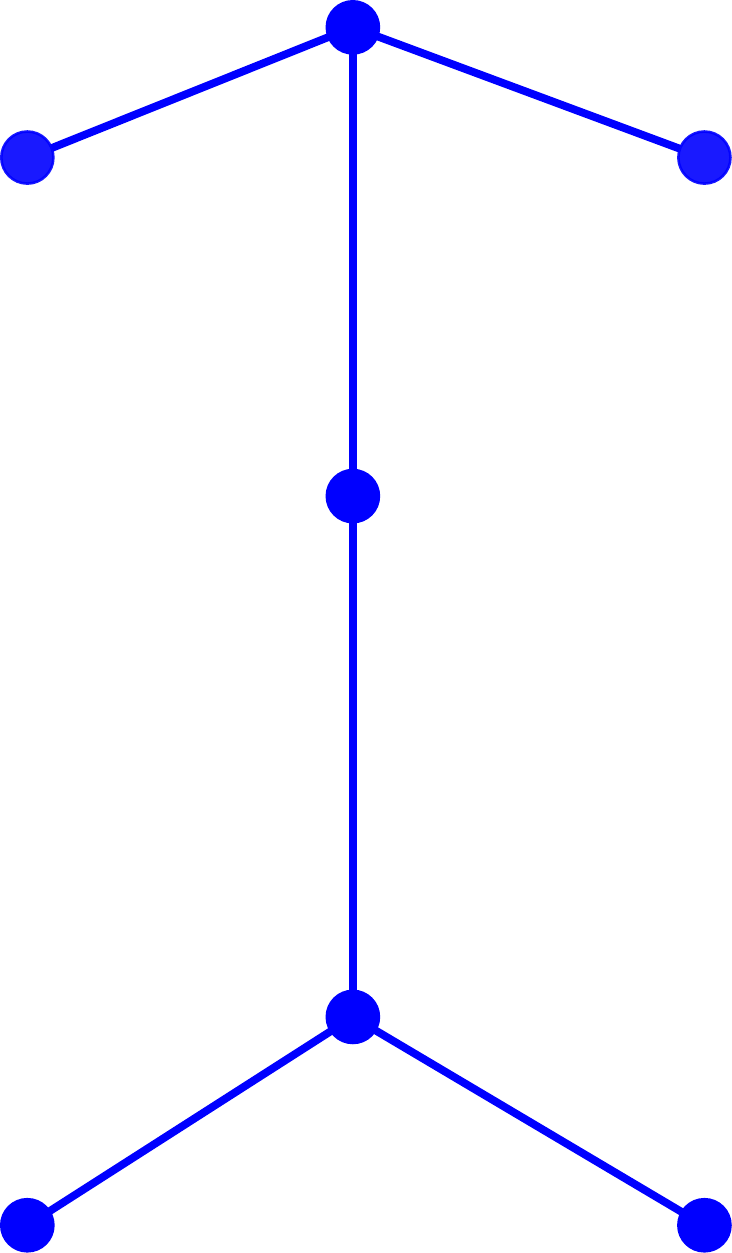} \\
            \footnotesize{Lower} & \footnotesize{Lower} \\
            \footnotesize{Appendicular} & \footnotesize{Axial} \\
        \end{tabular}
        &
        \begin{tabular}{ccc}
            \includegraphics[width=0.05\textwidth]{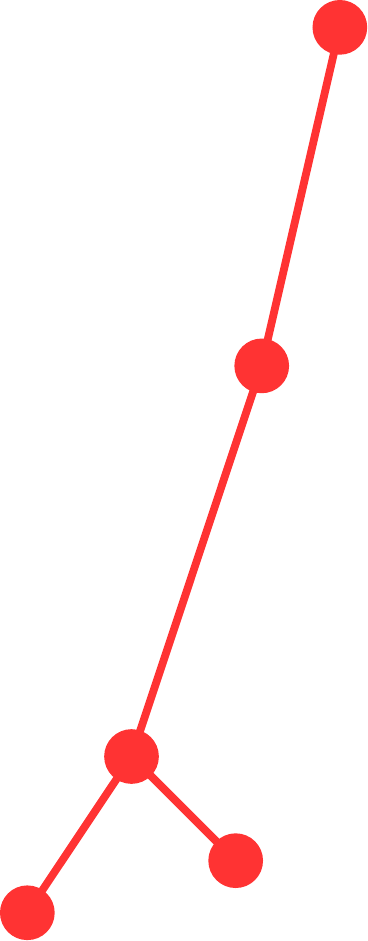} & 
            \includegraphics[width=0.10\textwidth]{images/camera_ready/upper_axial.pdf} &
            \includegraphics[width=0.05\textwidth]{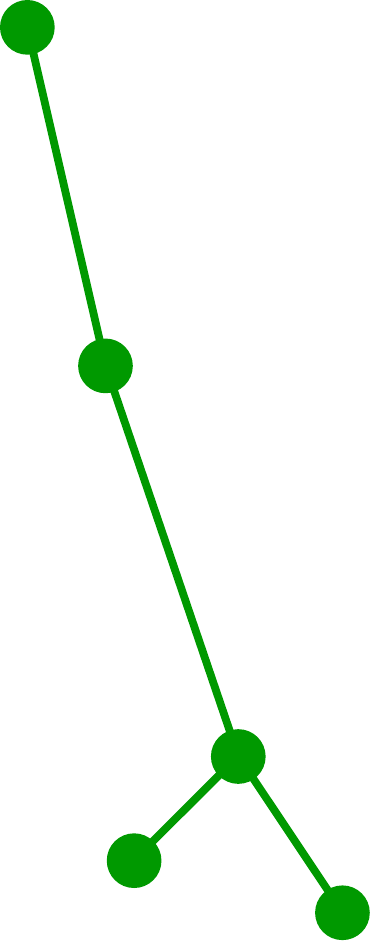} \\
            \footnotesize{Upper Left} & \footnotesize{Upper} & \footnotesize{Upper Right} \\
            \footnotesize{Appendicular} & \footnotesize{Axial} & \footnotesize{Appendicular} \\
            \includegraphics[width=0.05\textwidth]{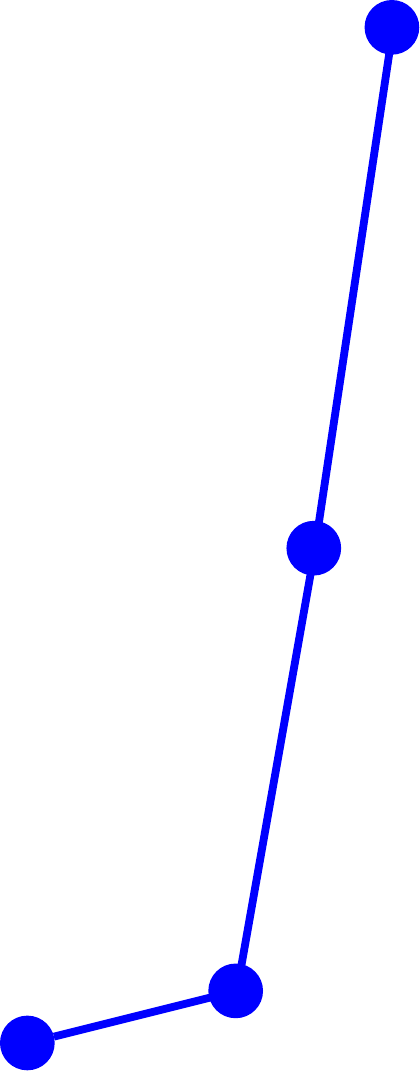} &
            \includegraphics[width=0.10\textwidth]{images/camera_ready/lower_axial.pdf} &
            \includegraphics[width=0.03\textwidth]{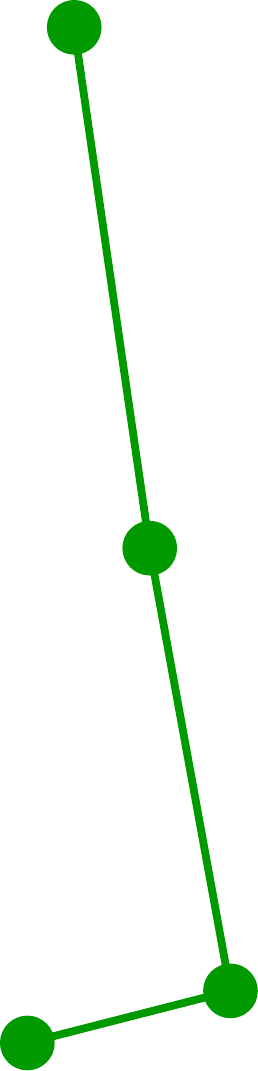} \\
            \footnotesize{Lower Left} & \footnotesize{Lower} & \footnotesize{Lower Right} \\
            \footnotesize{Appendicular} & \footnotesize{Axial} & \footnotesize{Appendicular} \\
        \end{tabular} \\
        \small{(c) Four parts} & \small{(d) Six parts}
    \end{tabular}
\end{center}
\caption{\small{\textbf{(a)} Geometric and Kinematic features, \textbf{(b)} Appendicular and axial body parts: two parts, \textbf{(c)} Dividing the appendicular and axial skeletons into upper and lower parts: four parts, \textbf{(d)} Dividing appendicular upper and lower skeletons into left and right: six parts.}}
\label{fig:overview}
\end{figure}

The major contributions of this paper are: (i) Formulation of a general part-based graph convolutional network (PB-GCN) which can be learned for any graph with well-known properties and its application to recognize actions from S-videos, (ii) Use of geometric and motion features in place of 3D joint locations at each vertex to boost recognition performance, and (iii) Exceeding the state-of-the-art on challenging benchmark datasets NTURGB+D and HDM05. The overview of our representation and signals is shown in Figure \ref{fig:overview}.

\section{Related Work}
\label{sec:relatedwork}
\subsection{Non graph-based methods}
\label{sec:2_1}
Skeletal action recognition has been approached using techniques such as handcrafted feature encodings, complex LSTM networks, image encodings with pretrained CNNs and non-euclidean methods based on manifolds. Non-deep learning methods worked well initially and proved usefulness of several extracted information from S-videos such as joint angles \cite{ofli2014sequence}, distances \cite{xia2012view} and kinematic features \cite{zanfir2013moving}. These methods learn from hand designed features using shallow models which do not model spatio-temporal properties of actions very well and constrain learning capacity.
\begin{figure}[t]
    \begin{center}
        \begin{tabular}{@{}ccc@{}}
            \includegraphics[width=0.3\textwidth]{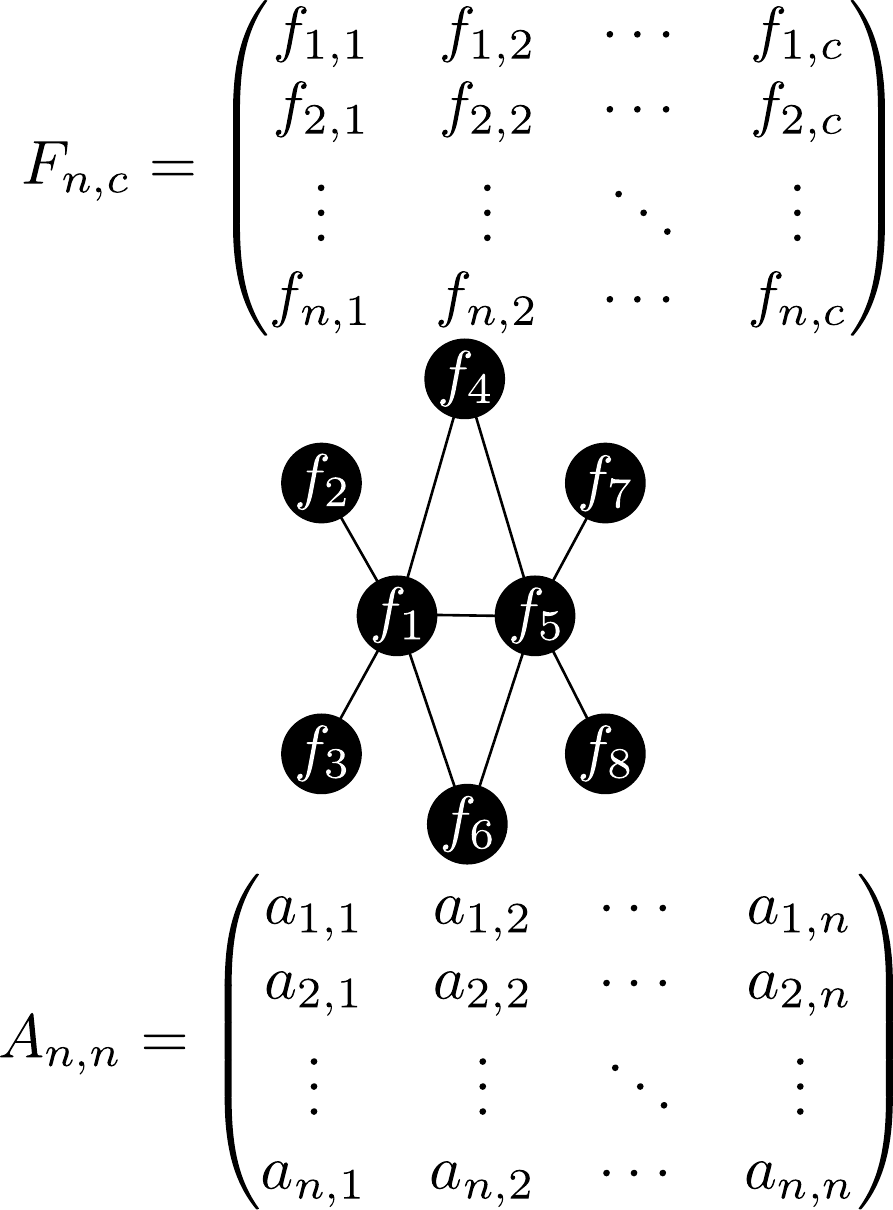} & 
            \includegraphics[width=0.3\textwidth]{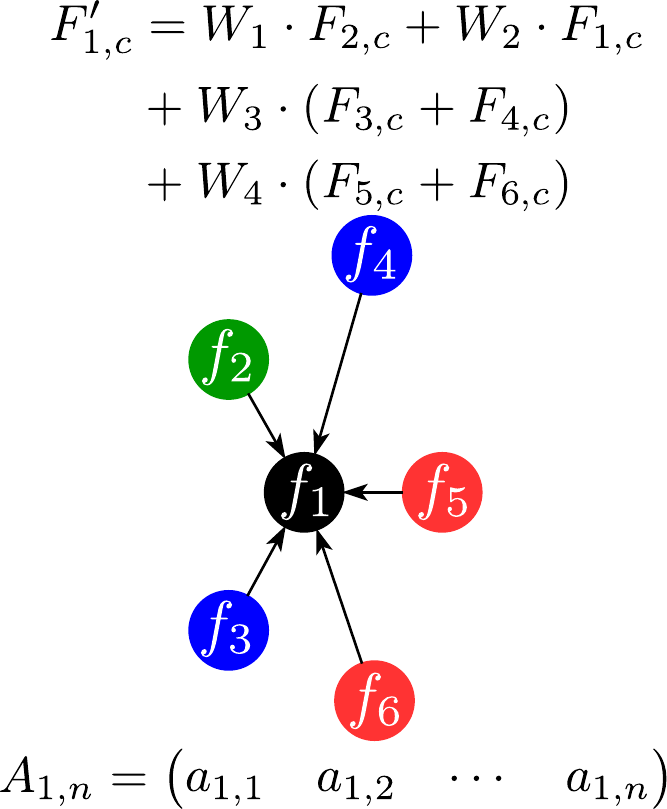} &
            \includegraphics[width=0.3\textwidth]{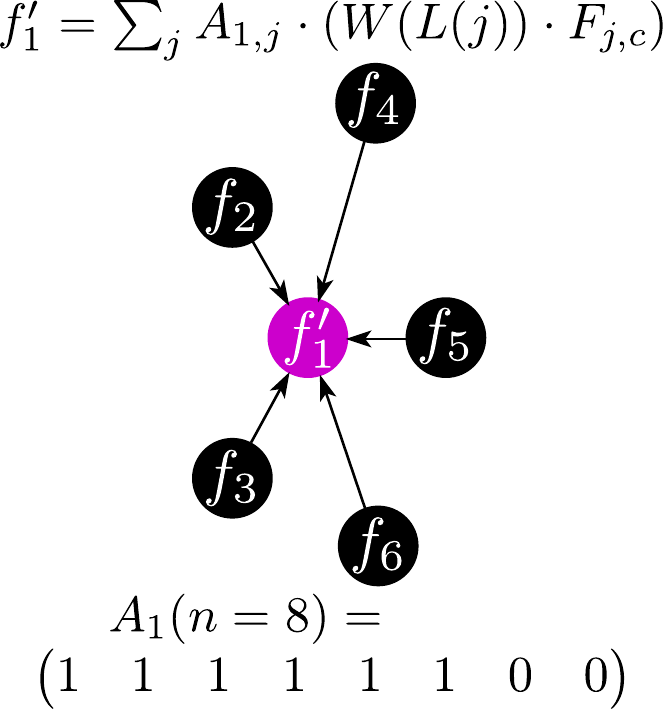} \\
            \footnotesize{(a) Graph feature} & \footnotesize{(b) Convolution for receptive field of} & \multirow{2}{*}{\footnotesize{(c) Final convolution equation}} \\
            \footnotesize{and adjacency matrices} & \footnotesize{a chosen root vertex $f_1$} & 
        \end{tabular}
    \end{center}
    \caption{\small{Equation-based formulation and illustration of a graph convolution}}
    \label{fig:gconv}
\end{figure}

On the other hand, LSTM-based methods were used because S-videos can be thought of as time sequences of features. Spatio-temporal LSTMs \cite{liu2016spatio, liu2017global}, attention-based LSTM \cite{song2017end} and simple LSTM networks with part-based skeleton representation \cite{tao2015moving, 7298714} have been used. These methods either use complex LSTM models which have to be trained very carefully or use part-based representation with a simple LSTM model. We propose a part-based graph convolutional network that has good learning capacity and uses a part-based representation, inheriting the good qualities of both types of aforementioned approaches. Image encodings of skeletons were proposed to facilitate usage of Imagenet pretrained CNNs to extract spatio-temporal features. Ke \etal \cite{ke2017new} generate images using relative coordinates while Du \etal \cite{7486569} and Li \etal \cite{li20183d} proposed a body part-based image encoding. Due to inherent differences in information in such image encodings and RGB images, it is almost impossible to interpret the learned filters. In contrast, our method is intuitive as it uses a graph-based representation for human skeleton.

Manifold learning techniques have been used for skeletal action recognition, where actions are represented as curves on Lie groups \cite{vemulapalli2014human} and Riemannian manifold \cite{devanne20153}. Deep learning on these manifolds is difficult \cite{huang2017deep} while deep learning on graphs (also a manifold) has developed recently \cite{defferrard2016convolutional, kipf2016semi}. Our method uses a human skeleton graph and learns a model using part-based graph convolutional network, exploiting the benefits of deep learning on graphs.

\subsection{Graph-based methods}
\label{sec:2_2}
Representing S-videos as skeleton graph sequences for recognizing actions had not been explored until recently. Li and Leung \cite{li2017graph} construct graphs using a statistical variance measure dependent on joint distances and match them for recognition. Recently, Yan \etal \cite{yan2018spatial} and Li \etal \cite{li2018spatio} proposed a spatio-temporal graph convolutional network for action recognition from S-videos. Both the methods construct graphs where the human skeleton is treated as a single graph. Our formulation explores a partitioned skeleton graph with a part-based graph convolutional network and we show that it improves recognition performance. Also, we use relative coordinates and temporal displacements as features at each vertex instead of 3D joint coordinates (see Figure \ref{fig:overview}(a)) which improves action recognition performance.

\section{Background}
\label{sec:background}
A graph is defined as $\mathcal{G} = (\mathcal{V}, \hspace{0.1cm} \mathcal{E})$ where $\mathcal{V}$ is the set of vertices and $\mathcal{E} \subseteq (\mathcal{V} \times \mathcal{V})$ is the set of edges. $\mathbf{A}$ is the graph adjacency matrix having $\mathbf{A}(i,j) = w, \hspace{0.1cm} w \in \mathbb{R} \setminus \{0\}$ if $(v_i, v_j) \in \mathcal{E}$ and $\mathbf{A}(i,j) = 0$ otherwise. $\mathcal{N}_k: v \rightarrow \mathcal{V}$ defines the set of vertices $\mathcal{V}$ in $k$-neighborhood of $v$ which includes neighbors having shortest path length atmost $k$ from vertex $v$. A labeling function $\mathbf{L}: \mathcal{V} \rightarrow \{0,1,\ldots,\mathcal{L}-1\}$ assigns a label to each vertex in a vertex set $\mathcal{V}$, where $\mathcal{L}$ is the number of unique labels. The adjacency matrix is normalized using a degree matrix as:
\begingroup
\small
\begin{align}
    \mathcal{D}(i,i) = \sum_{j} \mathbf{A}(i,j); \hspace{0.2cm} \mathbf{A}^{\mathbf{norm}} = \mathcal{D}^{-1/2}\mathbf{A}\mathcal{D}^{-1/2} \label{eq:normadj}
\end{align}
\endgroup
Graph convolutions can be formulated using spectral graph theory \cite{defferrard2016convolutional} or spatial convolution \cite{niepert2016learning} on graphs. We focus on spatial convolutions in this paper as they resemble convolutions on regular grid graphs like RGB images \cite{niepert2016learning}. A graph CNN can then be formed by stacking multiple graph convolution units. Graph convolution (shown in Figure \ref{fig:gconv}) can be defined as \cite{niepert2016learning}:
\begingroup
\small
\begin{align}
    \mathbf{Y}(v_i) &= \sum_{v_j \in \mathcal{N}_k(v_i)} \mathbf{W}(\mathbf{L}(v_j)) \mathbf{X}(v_j) \label{eq:npconv}
\end{align}
\endgroup
where, $v_i$ is the root vertex at which the convolution is centered (like center pixel in an image convolution), $\mathbf{W}(\cdot)$ is a filter weight vector of size of $\mathcal{L}$ indexed by the label assigned to neighbor $v_j$ in the $k$-neighborhood $\mathcal{N}_k(v_i)$, $\mathbf{X}(v_j)$ is the input feature at $v_j$ and $\mathbf{Y}(v_i)$ is the convolved output feature at root vertex $v_i$. Equation \ref{eq:npconv} can be written in terms of adjacency matrix as:
\begingroup
\small
\begin{align}
    \mathbf{Y}(v_i) &= \sum_{j} \hspace{0.1cm} \mathbf{A}^{\mathbf{norm}}(i,j) \hspace{0.1cm} \mathbf{W}(\mathbf{L}(v_j)) \hspace{0.1cm} \mathbf{X}(v_j) \label{eq:npconvadj}
\end{align}
\endgroup
$\mathbf{A}^{\mathbf{norm}}(i,j)$ basically defines the neighbors at distance $1$ and hence, Equation \ref{eq:npconv} captures a more general form of convolution by using $k$-order neighborhood $\mathcal{N}_k(v_i)$.

\begin{figure}[t]
\begin{center}
    \begin{tabular}{@{}c@{}c@{}c@{}}
        \begin{tabular}{cc}
            \includegraphics[width=0.12\textwidth]{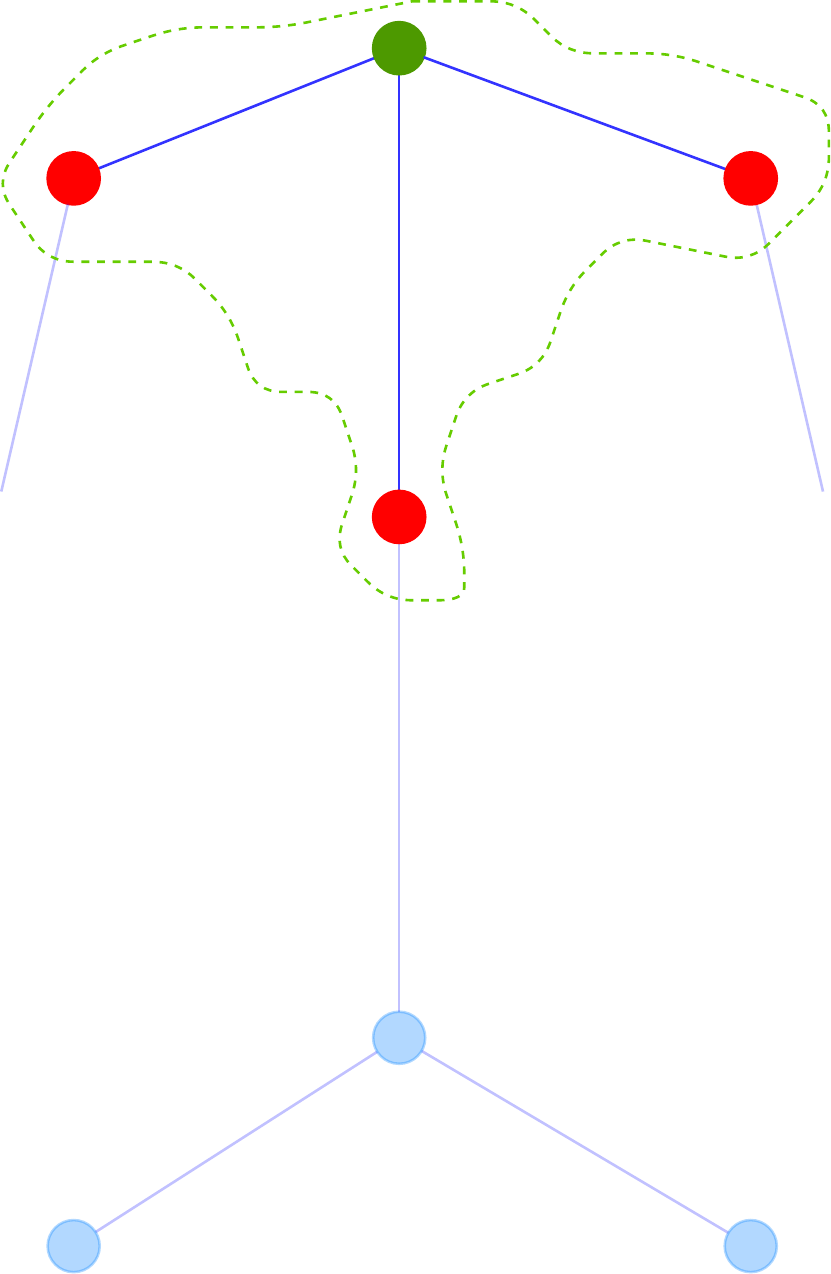} &
            \includegraphics[width=0.12\textwidth]{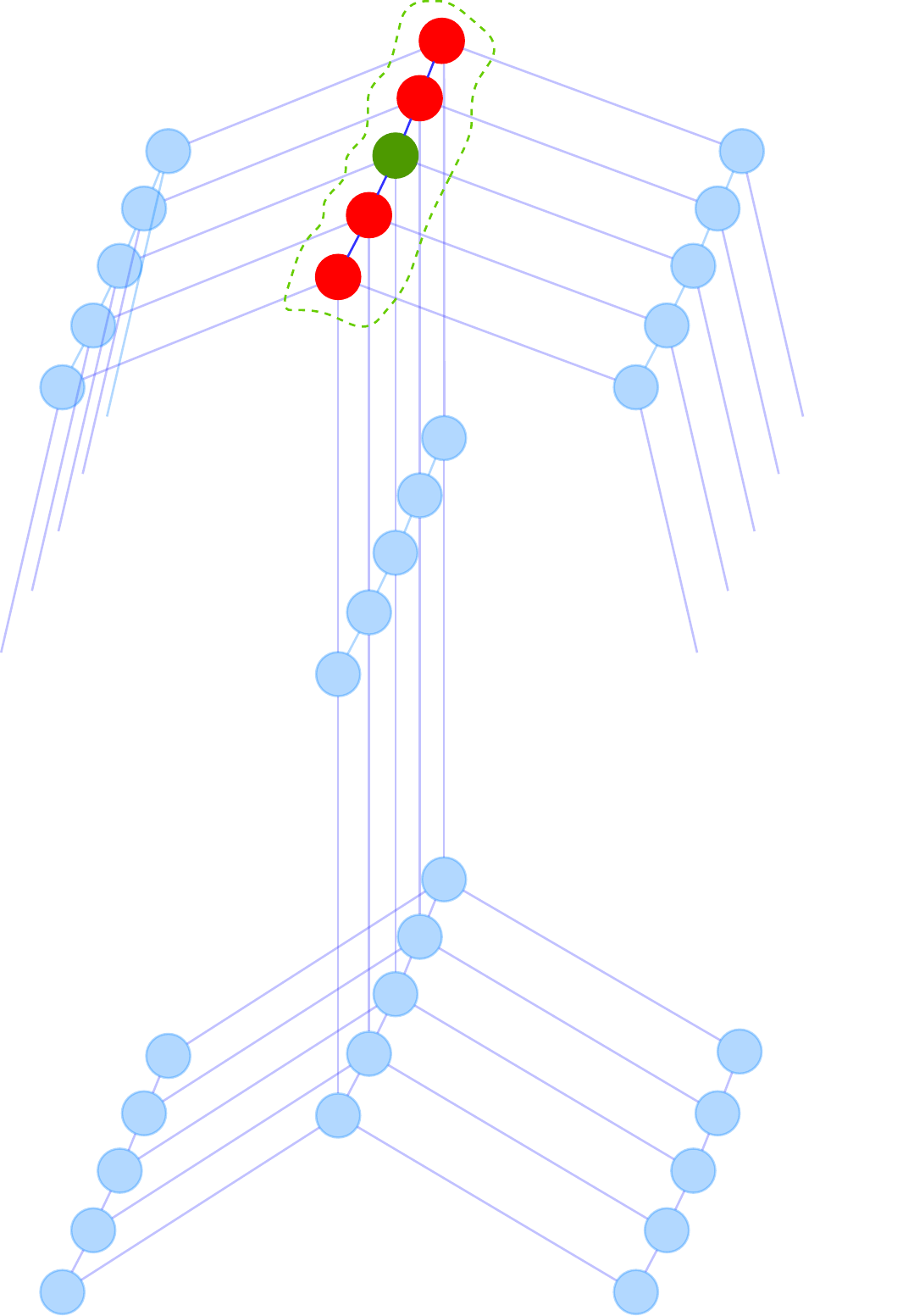} \\
            \footnotesize{Spatial neighbours} & \footnotesize{Temporal neighbours}
        \end{tabular} &
        \begin{tabular}{c}
            \includegraphics[width=0.15\textwidth]{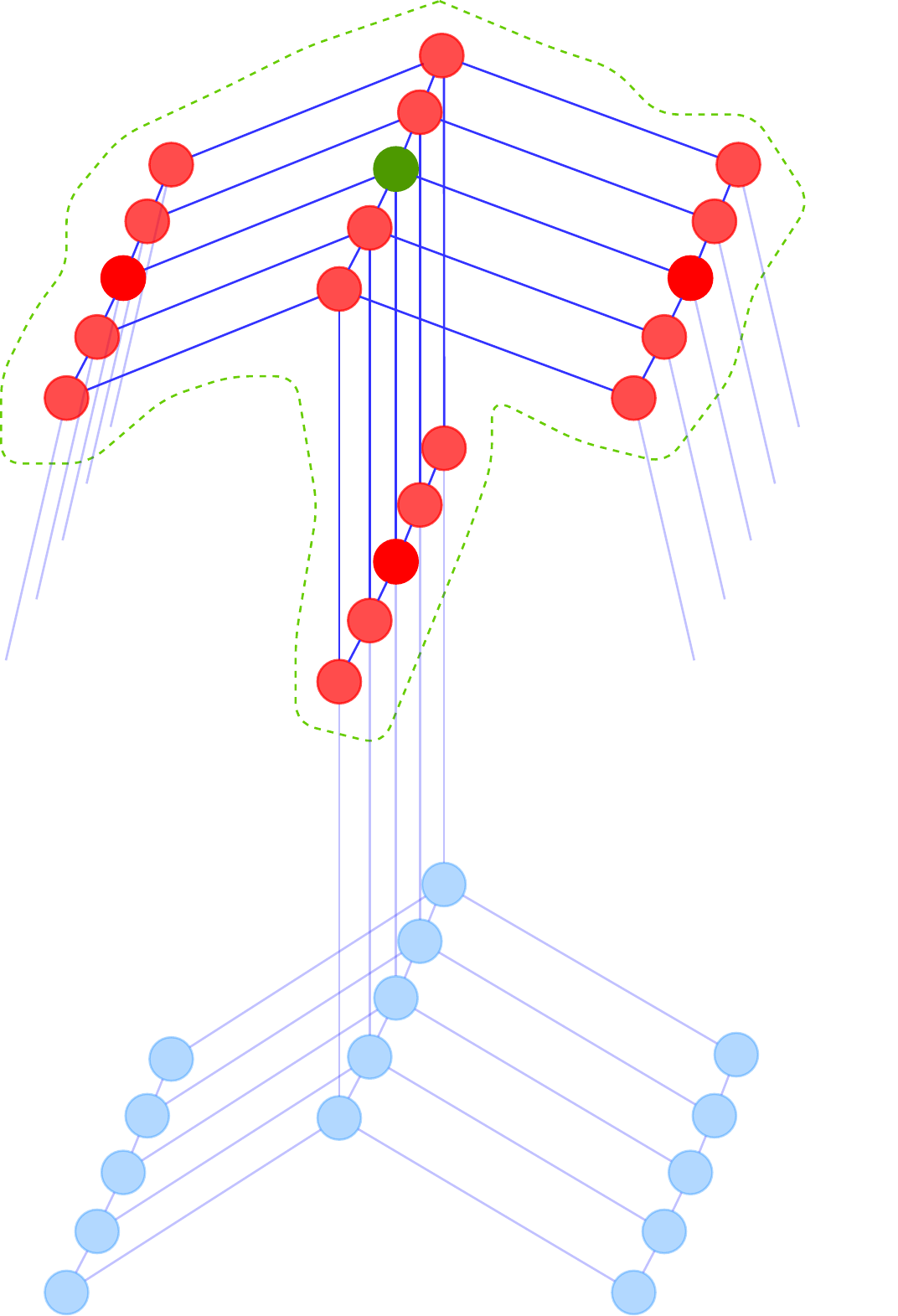} \\
            \footnotesize{Spatio-temporal neighbours}
        \end{tabular} &
        \begin{tabular}{c}
            \includegraphics[width=0.15\textwidth]{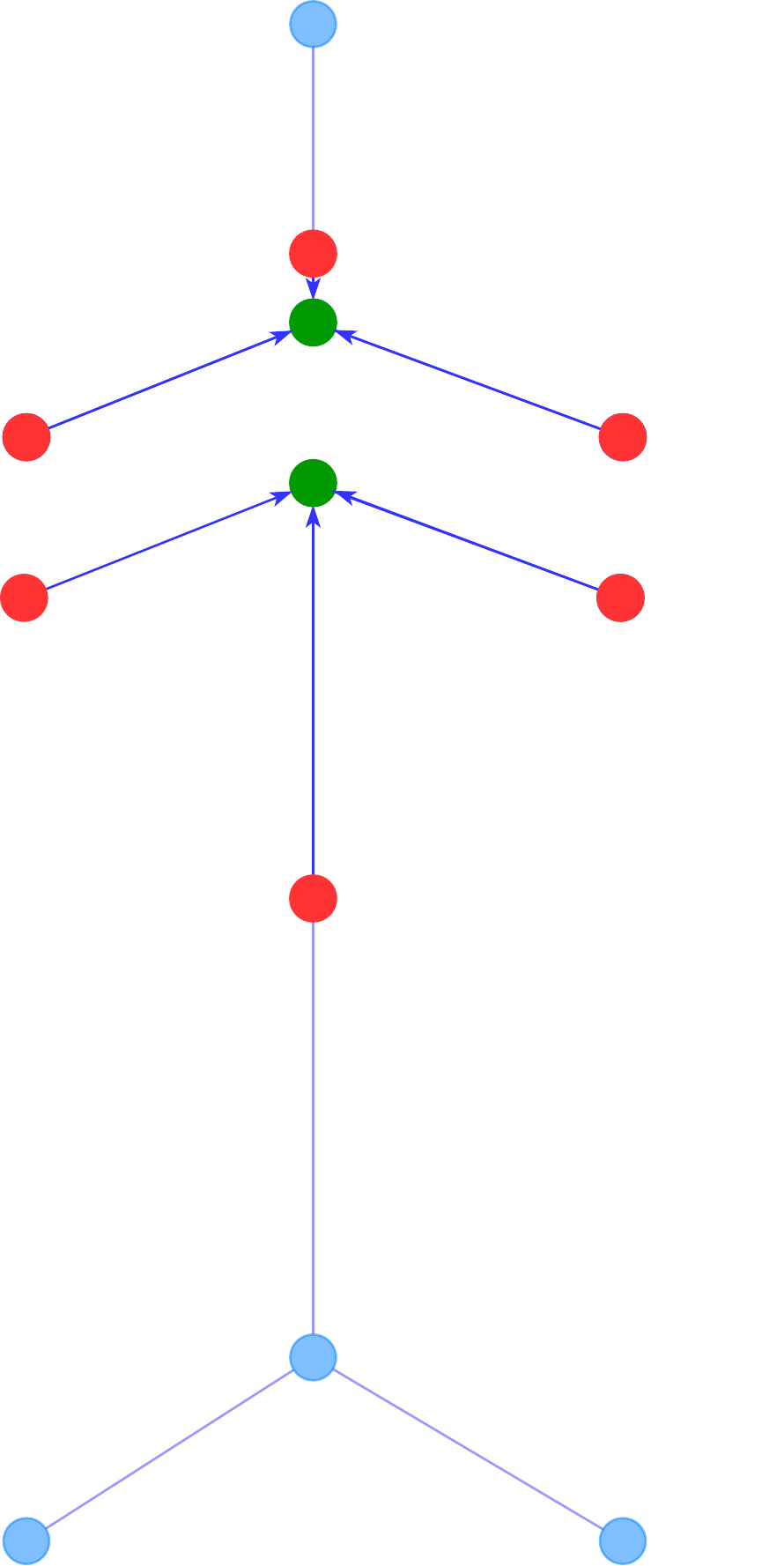} \\
            \footnotesize{Single spatial convolution}
        \end{tabular} \\
        \small{(a)} & \small{(b)} & \small{(c)} \\
        \begin{tabular}{c}
            \includegraphics[width=0.15\textwidth]{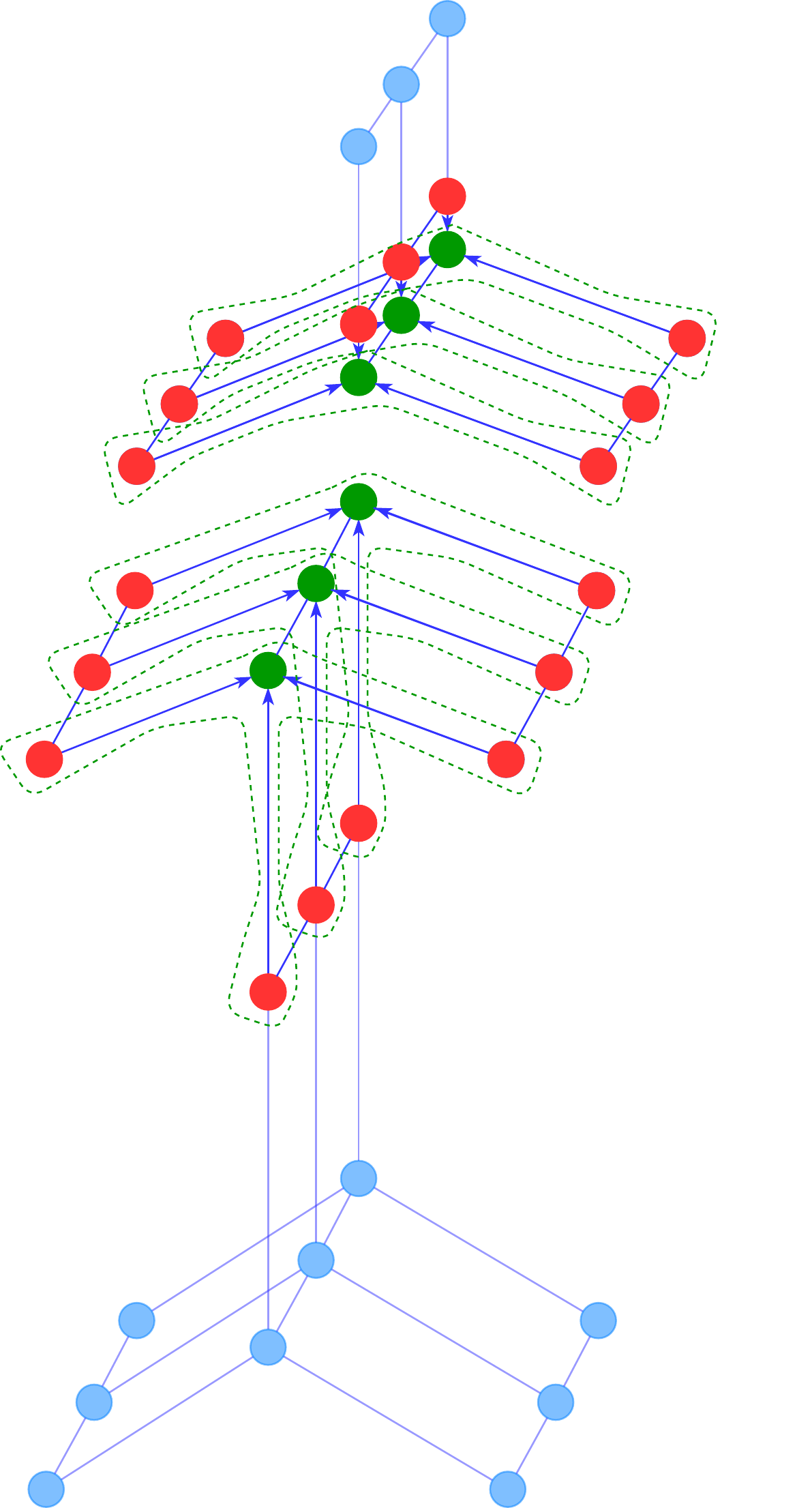} \\
            \footnotesize{Multiple spatial convolutions} \\
            \footnotesize{across time}
        \end{tabular} &
        \begin{tabular}{c}
            \includegraphics[width=0.15\textwidth]{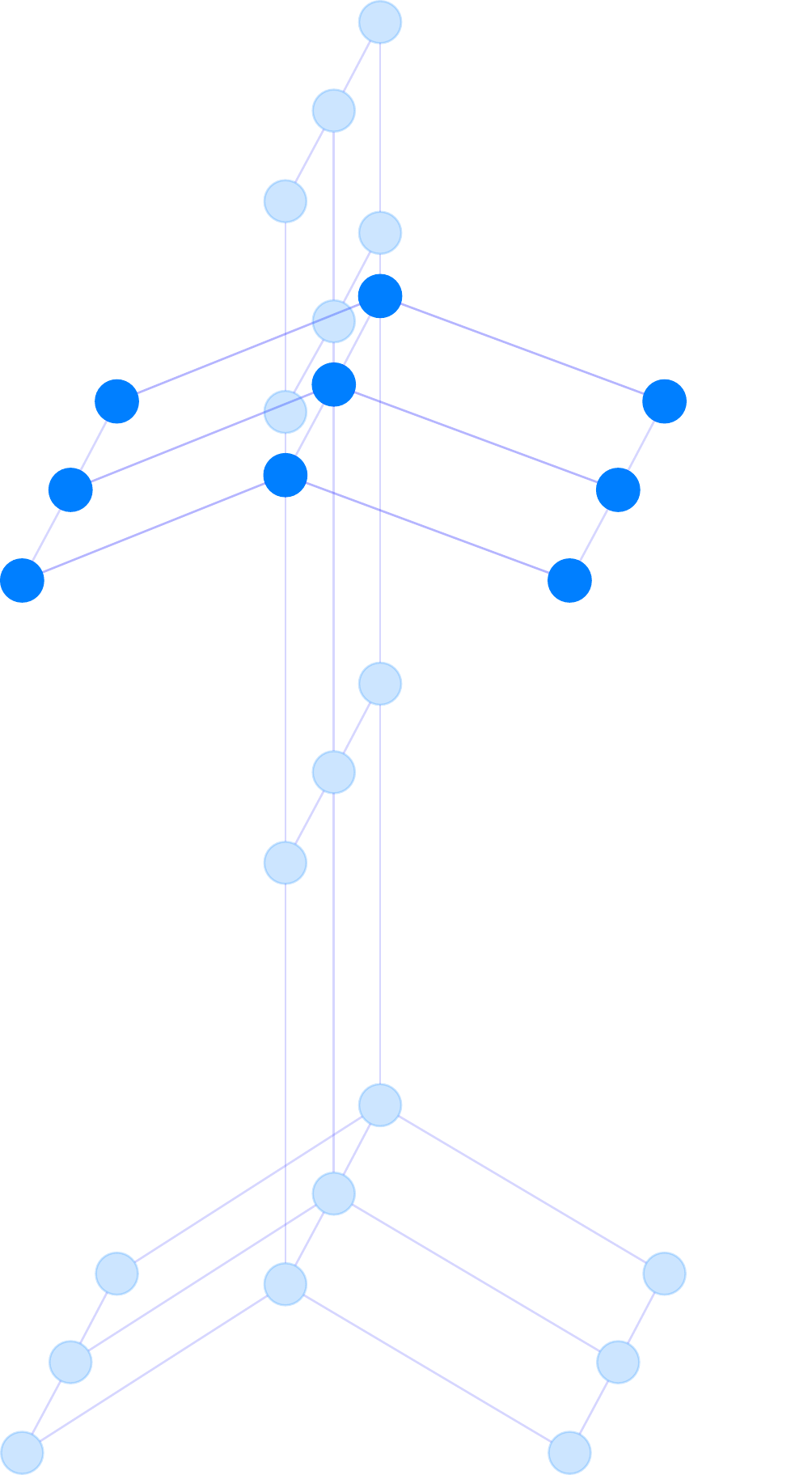} \\
            \footnotesize{Combination of Head and} \\
            \footnotesize{Torso parts: $\mathcal{F}_{agg}$}
        \end{tabular} &
        \begin{tabular}{c}
            \includegraphics[width=0.15\textwidth]{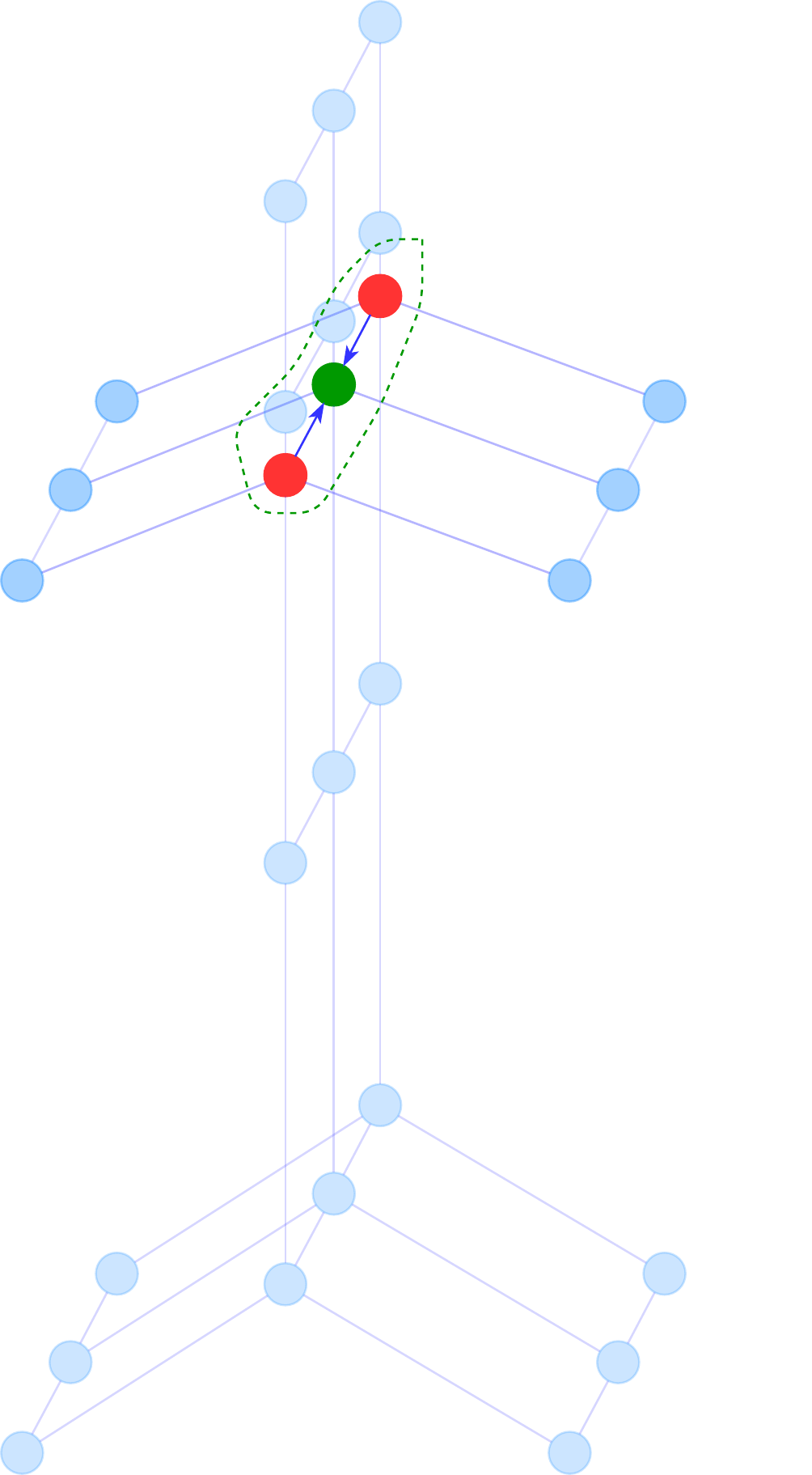} \\
            \footnotesize{Temporal convolution after} \\
            \footnotesize{$\mathcal{F}_{agg}$ is applied}
        \end{tabular} \\
        \small{(d)} & \small{(e)} & \small{(f)}
    \end{tabular}
\end{center}
\caption{\small{Spatio-temporal neighborhood for root node (in green) and depiction of convolutions in space and time dimensions. Effect of application of $\mathcal{F}_{agg}$ is shown, where the common vertices are in darker shade.}}
\label{fig:pstgcn}
\end{figure}

\subsection{Part-based Graph}
\label{sec:3_1}
Graphs representing real world manifolds can often be thought of as being made up of several parts. For instance, a graph representing a complex molecule consists of several simple structures, such as structure of a protein biomolecule, which can be divided into several polypeptide chains that make up the complex. Similarly, human body can be visualized as connected rigid parts, much like a deformable part-based model \cite{Felzenszwalb:2005}. The graph of the skeleton of human body can be divided into parts, where each subgraph represents a part of the human body.

In general, a part-based graph can be constructed as a combination of subgraphs where each subgraph has certain properties that define it. Let us consider that a graph $\mathcal{G}$ has been divided into $n$ partitions. Formally:
\begingroup
\small
\begin{align}
    \mathcal{G} = \bigcup_{p \in \{1,\ldots,n\}} \mathcal{P}_p \hspace{0.1cm} | \hspace{0.1cm} \mathcal{P}_p = (\mathcal{V}_p, \mathcal{E}_p) \label{eq:partgraph}
\end{align}
\endgroup
$\mathcal{P}_p$ is the partition (or subgraph) $p$ of the graph $\mathcal{G}$. We consider scenarios in which the partitions can share vertices or have edges connecting them. We proceed to explain how the part-based graph convolution is defined for the part-based graph.

\subsection{Part-based Graph Convolutions}
\label{sec:3_2}
In essence, graph convolutions over parts are aimed at capturing high-level properties of parts and learn the relations between them. In a Deformable Part-based Model, different parts are identified and relations between them are learned through the deformation of the connections between them. Similarly, graph convolutions over a part identifies the properties of that subgraph and an aggregation across subgraphs learns the relations between them. For a part-based graph, convolutions for each part are performed separately and the results are combined using an aggregation function $\mathcal{F}_{agg}$. Using $\mathcal{F}_{agg}$ over edges across partitions:
\begingroup
\small
\begin{align}
    \mathbf{Y}_p(v_i) &= \sum_{v_j \in \mathcal{N}_{kp}(v_i)} \mathbf{W}_p(\mathbf{L}_p(v_j)) \mathbf{X}_p(v_j), \hspace{0.1cm} p \in \{1,\ldots,n\} \label{eq:pconv}\\
    \mathbf{Y}(v_i) &= \mathcal{F}_{agg}(\mathbf{Y}_{p1}(v_i), \mathbf{Y}_{p2}(v_j)) \hspace{0.1cm} | \hspace{0.1cm} (v_i, v_j) \in \mathcal{E}_{(p1, p2)}, \hspace{0.1cm} (p1, p2) \in \{1,\ldots,n\} \times \{1,\ldots,n\} \label{eq:peagg}
\end{align}
\endgroup
Using $\mathcal{F}_{agg}$ for common vertices across partitions:
\begingroup
\small
\begin{align}
    \mathbf{Y}(v_i) &= \mathcal{F}_{agg}(\mathbf{Y}_{p1}(v_i), \mathbf{Y}_{p2}(v_i)) \hspace{0.1cm} | \hspace{0.1cm} (p1, p2) \in \{1,\ldots,n\} \times \{1,\ldots,n\} \label{eq:pvagg}
\end{align}
\endgroup
The convolution parameters $\mathbf{W}_p$ can be shared across parts or kept separate, while the neighbors of $v_i$ only in that part $(\mathcal{N}_{kp}(v_i))$ are considered. In order to combine the information across parts, the function $\mathcal{F}_{agg}$ combines information at shared vertices (equation \ref{eq:pvagg}) or shares information through edges crossing parts (equation \ref{eq:peagg}, $\mathcal{E}_{(p1,p2)}$ contains all edges connecting parts p1 and p2), according to the partition configuration. A sophisticated $\mathcal{F}_{agg}$ can be employed to make the model powerful. Using graph convolutions, part-based graph models can learn rich representations and we demonstrate the strength of this model through application to action recognition from S-videos.

\section{Spatio-temporal Part-based Graph Convolutions}
\label{sec:stmodel}
The S-videos are represented as spatio-temporal graphs. In order to include the temporal dimension, corresponding joints in each part are connected temporally. Figure \ref{fig:pstgcn}(b) shows the spatio-temporal graph for \textit{torso} over \textit{five} frames. Adapting select-assemble-normalize (\textsc{patchy-san}) proposed by Niepert \etal \cite{niepert2016learning} we present an overview of convolution formulation for our spatio-temporal graph by extending ideas from section \ref{sec:3_2}. For in-depth understanding, we refer the reader to \cite{niepert2016learning}. We perform a spatial convolution on each partition following equation \ref{eq:pconv}, combine the convolved partitions using $\mathcal{F}_{agg}$ and perform temporal convolution on the graph obtained by aggregating the partitions. In effect, we spatially convolve each partition independently for each frame, aggregate them at each frame and perform temporal convolution on the temporal dimension of the aggregated graph. For a possible partitioning of human skeleton, this phenomenon is shown in Figure \ref{fig:pstgcn}(c) for spatial convolution for a vertex common to torso and head, \ref{fig:pstgcn}(d) for spatial convolutions in different frames, \ref{fig:pstgcn}(e) for applying $\mathcal{F}_{agg}$ on head + torso and \ref{fig:pstgcn}(f) for convolution on temporal dimension of the combined graph.

We first define the spatial and temporal neighborhood of a vertex in spatio-temporal graph and assign labels to the vertices in the neighborhoods, which is required to perform convolutions. For each vertex, we use 1-neighborhood $(k = 1)$ for spatial dimension $(\mathcal{N}_1)$ as the skeleton graph is not very large and a $\tau$-neighborhood $(k = \tau)$ for the temporal dimension $(\mathcal{N}_{\tau})$. Figure \ref{fig:pstgcn}(a) (dashed polygons) shows the spatial \& temporal neighborhood for a \textbf{root} vertex. The different neighborhood sets for our model are defined as ($\mathbf{d}(v_i, v_j)$ = length of shortest path between $v_i$ and $v_j$):
\begingroup
\small
\begin{align}
    \mathcal{N}_{1p}(v_i) &= \{v_j \hspace{0.1cm} | \hspace{0.1cm} \mathbf{d}(v_i, v_j) \hspace{0.1cm} \leq \hspace{0.1cm} 1, \hspace{0.1cm} v_i, v_j \in \mathcal{V}_p\} \label{eq:sneigh} \\
    \mathcal{N}_{\tau}(v_{i{t_a}}) &= \{v_{i{t_b}} \hspace{0.1cm} | \hspace{0.1cm} \mathbf{d}(v_{i{t_a}}, v_{i{t_b}}) \hspace{0.1cm} \leq \hspace{0.1cm} \left\lfloor \frac{\tau}{2} \right\rfloor\} \label{eq:tneigh}
\end{align}
\endgroup
where, $t_a \hspace{0.1cm} \& \hspace{0.1cm} t_b$ represent two time instants and $p \in \{1,\ldots,n\}$ is the partition index. The set of vertices $\mathcal{V}_p$ differs for each part, with some vertices shared between parts (Figure \ref{fig:overview}(c)). As temporal convolution is performed on the aggregated spatio-temporal graph, $\mathcal{N}_{\tau}$ is not part-specific. Figure \ref{fig:pstgcn}(a) shows the spatial and temporal neighborhoods for a \textbf{root} vertex in \textit{torso}. For ordering vertices in the receptive fields (or neighborhoods), we use a single label spatially $(\mathbf{L}_S: \mathcal{V} \rightarrow \{0\})$ to weigh vertices in $\mathcal{N}_{1p}$ of each vertex equally and $\tau$ labels temporally $(\mathbf{L}_T: \mathcal{V} \rightarrow \{0,\ldots,\tau-1\})$ to weigh vertices across frames in $\mathcal{N}_{\tau}$ differently. The labeling functions are defined as:
\begingroup
\small
\begin{align}
    \mathbf{L}_{S}(v_{jt}) &= \{0 \hspace{0.1cm} | \hspace{0.1cm} v_{jt} \in \mathcal{N}_{1p}(v_{it})\} \label{eq:slabel} \\
    \mathbf{L}_{T}(v_{i{t_b}}) &= \{((t_b - t_a) + \left\lfloor \frac{\tau}{2} \right\rfloor) \hspace{0.1cm} | \hspace{0.1cm} v_{i{t_b}} \in \mathcal{N}_{\tau}(v_{i{t_a}})\} \label{eq:tlabel}
\end{align}
\endgroup
Using the labeled spatial and temporal receptive fields, we define the spatial and temporal convolutions as (adapted from \cite{kipf2016semi}):
\begingroup
\small
\begin{align}
    \mathbf{Y}_p(v_{it}) &= \sum_{v_{jt} \hspace{0.02cm} \in \hspace{0.02cm} \mathcal{N}_{1p}(v_{it})}\mathbf{A}_p(i, j) \hspace{0.05cm} \mathbf{Z}_p(v_{jt}) \hspace{0.1cm} | \hspace{0.1cm} p \in \{1,\ldots,n\} \label{eq:sconvpart} \\
    \mathbf{Z}_p(v_{jt}) &= \mathbf{W}_{p}(\mathbf{L}_{S}(v_{jt})) \hspace{0.1cm} \mathbf{X}_p(v_{jt}) \label{eq:xtoz} \\
    \mathbf{Y}_{S}(v_{it}) &= \mathcal{F}_{agg}(\{\mathbf{Y}_1(v_{it}), \ldots, \mathbf{Y}_n(v_{it})\}) \label{eq:sconvfagg} \\
    \mathbf{Y}_{T}(v_{i{t_a}}) &= \sum_{v_{j{t_b}} \hspace{0.02cm} \in \hspace{0.02cm} \mathcal{N}_{\tau}(v_{i{t_a}})} \mathbf{W}_{T}(\mathbf{L}_{T}(v_{i{t_b}})) \hspace{0.05cm} \mathbf{Y}_{S}(v_{i{t_b}}) \label{eq:tconv}
\end{align}
\endgroup
where, $\mathbf{A}_p$ is a normalized adjacency matrix as explained in section \ref{sec:background} for part $p$. $\mathbf{L}_{S}$ for each part is same but $\mathcal{N}_{1p}$ is part-specific. $\mathbf{W}_{p} \in \mathbb{R}^{C^{\prime} \times C \times 1 \times 1}$ is a part-specific channel transform kernel (pointwise operation) and $\mathbf{W}_{T} \in \mathbb{R}^{C^{\prime} \times C^{\prime} \times \tau \times 1}$ is the temporal convolution kernel. $\mathbf{Z}_p$ is the output from applying $\mathbf{W}_p$ on input features $\mathbf{X}_p$ at each vertex. $\mathbf{Y}_{S}$ is the output obtained after aggregating all partition graphs at one frame and $\mathbf{Y}_{T}$ is the output after applying temporal convolution on $\mathbf{Y}_{S}$ output of $\tau$ frames. We use a weighted sum fusion as our $\mathcal{F}_{agg}$:
\begingroup
\small
\begin{align}
    \mathcal{F}_{agg}(\{\mathbf{Y}_1,\ldots,\mathbf{Y}_n\}) &= \sum_{i} \mathbf{W}_{agg}(i) \hspace{0.05cm} \mathbf{Y}_i \label{eq:fagg}
\end{align}
\endgroup

Human skeleton can be divided into two major components: (1) Axial skeleton and (2) Appendicular skeleton. The body parts included in these two components are shown in Figure \ref{fig:overview}(b). Human skeleton can be divided into parts based on these components. Different division schemes are shown in Figure \ref{fig:overview}(b), \ref{fig:overview}(c) and \ref{fig:overview}(d) and we use these schemes for experiments to test our PB-GCN.

For the final representation, we divide the human skeleton into \textit{four} parts: \textbf{head}, \textbf{hands}, \textbf{torso} and \textbf{legs}, which corresponds to a division scheme where each of the axial and appendicular skeleton are divided into upper and lower components, as illustrated in Figure \ref{fig:overview}(c). We consider left and right parts of hands and legs together in order to be agnostic to \textit{laterality} \cite{lateral} (handedness / footedness) of the human when performing an action. To show how being agnostic to laterality is helpful, we divide the upper and lower components of appendicular skeleton into left and right (shown in Figure \ref{fig:overview}(d)), resulting in six parts and show results on it. To cover all natural connections between joints in skeleton graph, we include an overlap of atleast one joint between two adjacent parts. For example, in Figure \ref{fig:overview}(c), shoulder joints are common between the head and hands. For the lower appendicular skeleton (viz. legs), we also include the joint at the base of spine to get a good overlap with lower axial skeleton.

\paragraph{Architecture and Implementation} We represent each subgraph by its adjacency matrix, normalized by corresponding degree matrix $\mathcal{D}$. Our model takes as input a tensor having features for each vertex in the spatio-temporal graph of S-video and outputs a vector of class scores for the video. The architecture of the graph convolutional network is similar to Yan \etal \cite{yan2018spatial} and consists of $9$ spatio-temporal graph convolution units (each unit with the \textit{four} $\mathbf{W}_{p}$ kernels, \textit{one} $\mathbf{W}_{T}$ kernel and a residual) with an initial spatio-temporal head unit, based on a Resnet-like model \cite{he2016deep}. First three layers have 64 output channels, next three have 128 and last three have 256. We also use a learnable edge weight mask for learning edge weights in each subgraph \cite{yan2018spatial}. We use the Pytorch framework \cite{paszke2017automatic} for our implementation. The code and models are made publicly available: \href{https://github.com/dracarys983/pb-gcn}{\texttt{https://github.com/dracarys983/pb-gcn}}.

\section{Geometric \& Kinematic Signals}
\label{sec:signals}
Yan \etal \cite{yan2018spatial} use the 3D coordinates of each joint directly as the signal at each graph node. Relative coordinates \cite{zhang2017geometric, ke2017new} and temporal displacements \cite{zanfir2013moving} of joints have been used earlier for action recognition. Derived information like optical flow and Manhattan line map has been found useful on RGB images also \cite{wang2016temporal,zou2018layoutnet}. Even a CNN framework can be more effective and efficient if relevant derived information is supplied as input to the network.

We use a signal at each node that combines temporal displacements across time and relative coordinates, with respect to shoulders and hips \cite{ke2017new}.  This representation provides translation invariance to the representation \cite{verma2018feastnet} and improves skeletal action recognition performance significantly. Figure \ref{fig:overview}(a) illustrates the computation of the two signals for a single skeleton video frame. We show the effect of relative joint coordinates (geometric signal) and temporal displacements (kinematic signal) individually and the performance improvement obtained by using a combination of these signals for a baseline one-part model as well as our four part-based model in the Table \ref{tab:ablation}(b). The improvement in performance obtained using the geometric and kinematic signals is noteworthy.

\section{Experimental Setup and Results}
\label{sec:exp_res}
We use SGD as the optimizer and run the training for 80 epochs (NTURGB+D) / 120 epochs (HDM05). We set the initial learning rate to 0.1 and all the experiments are run on a cluster with $4$ Nvidia GTX 1080Ti GPUs. The batch size is set to 64. Learning rate decay schedule (set to decay by 0.1 at epochs 20, 50 and 70 for NTURGB+D, and at epoch 80 for HDM05) is finalized using a validation set. No augmentation is performed for any of the experiments, consistent with graph-based method \cite{yan2018spatial}. We perform ablation studies on the large-scale NTURGB+D dataset (shown in Table \ref{tab:ablation}) and then compare with state-of-the-art on both HDM05 and NTURGB+D using the best configuration of our model (shown in Table \ref{tab:sota}).

\begin{table}[t]
    \begin{center}
        \small
        \begin{tabular}{cc}
            \multirow{2}{*}{(a) Performance with number of parts} & (b) Performance with various signals for \\
            & best \& worst number of parts \\
            \begin{tabular}{ccc}
            \toprule
            \multirow{2}{*}{\textbf{\#Parts}} & \multicolumn{2}{c}{\textbf{Accuracy}} \\
            & CS & CV \\
            \midrule
            One & 79.4 & 87.9 \\
            Two & 80.2 & 88.4 \\
            Four & \textbf{82.8} & \textbf{90.3} \\
            Six & 81.4 & 89.1 \\
            \bottomrule
            \end{tabular}
            &
            \begin{tabular}{ccccc}
                \toprule
                \multirow{3}{*}{\textbf{Signals}} & \multicolumn{4}{c}{\textbf{Accuracy}} \\
                & \multicolumn{2}{c}{\textbf{\#Parts=1}} & \multicolumn{2}{c}{\textbf{\#Parts=4}} \\
                & CS & CV & CS & CV \\
                \midrule
                $J_{loc}$ & 79.4 & 87.9 & 82.8 & 90.3 \\
                $\mathbf{D}_{R}$ & 83.6 & 87.7 & 84.6 & 88.4 \\
                $\mathbf{D}_{T}$ & 84.3 & 91.6 & 85.4 & 92.6 \\
                $\mathbf{D}_{R} || \mathbf{D}_{T}$ & \textbf{85.6} & \textbf{91.8} & \textbf{87.5} & \textbf{93.2} \\
                \bottomrule
            \end{tabular}
        \end{tabular}
    \end{center}
    \caption{\small{Performance comparison for different number of parts in the skeleton graph and signals at vertices using our PB-GCN, on NTURGB+D \cite{Shahroudy_2016_CVPR} (CS: Cross Subject, CV: Cross View). The symbols for signals, $J_{loc}$: Absolute 3D joint locations, $\mathbf{D}_{R}$: Relative coordinates, $\mathbf{D}_{T}$: Temporal displacements and $\mathbf{D}_{R} || \mathbf{D}_{T}$: Concatenation of $\mathbf{D}_{R}$ and $\mathbf{D}_{T}$.}}
    \label{tab:ablation}
\end{table}

\subsection{Datasets}
\label{sec:5_1}
\paragraph{NTURGB+D}\cite{Shahroudy_2016_CVPR}
This is currently the largest RGBD dataset for action recognition to the best of our knowledge. It has 56,880 video sequences shot with three Microsoft Kinect v2 cameras from different viewing angles. There are 60 classes among the action sequences and 3D coordinates of 25 joints are provided for each human skeleton tracked. There is a large variation in viewpoint, intra-class subjects and sequence lengths, which makes this dataset challenging. We remove 302 of the captured samples having missing or incomplete skeleton data. The protocol mentioned in Shahroudy \etal \cite{Shahroudy_2016_CVPR} is followed for comparisons with previous methods.

\paragraph{HDM05}\cite{Muller07documentationmocap}
This dataset was captured by using an optical marker-based Vicon system. It contains 2337 action sequences ranging across 130 motion classes performed by \textit{five} actors. This dataset currently has the largest number of motion classes. The actors are named ``bd'', ``bk'', ``dg'', ``mm'' and ``tr'', and 31 joints are annotated for each skeleton. This dataset is challenging due to intra-class variations induced by multiple realizations of same action and large number of motion classes. We follow the protocol given in \cite{huang2017riemannian} which is used by recent deep learning methods.

\subsection{Discussion}
\label{sec:5_2}
\paragraph{\textsc{Part-based graph model}:} Our motivation to use a part-based graph model is derived primarily from the fact that human actions are made up of ``gestures'' which represent motion of a body part. The seminal success of DPMs \cite{Felzenszwalb:2005} in detecting humans in images reinforces the motivation further. We discuss the effect of proposed spatio-temporal part-based graph model below.

\paragraph{(a) How many parts to have?} We start with a coarse-grained scheme where entire skeleton is a single part and progress towards finer representations. The different partitions are, \textit{two parts}: dividing skeleton into axial and appendicular skeleton, \textit{four parts}: as explained in section \ref{sec:stmodel} and \textit{six parts}: Assigning left and right in hands and legs. The feature at each vertex in the input is 3D coordinate of the corresponding joint. From Table \ref{tab:ablation}(a), we can see that using two parts improves over one and four improves over two. This shows that partitioning the skeleton graph into subgraphs with useful properties helps. However, dividing upper and lower skeletons into left and right in four part scheme does not improve performance, as per our intuition about \textit{laterality} mentioned in section \ref{sec:stmodel}. This experiment suggests that part-based model improves performance over single part and being agnostic to laterality is helpful. Our final model uses the \textit{four} part division of the human skeleton.

\paragraph{(b) Comparison to graph-based models} From Table \ref{tab:sota}(a) and Table \ref{tab:ablation}(b), it can be seen that our part-based model performs better than graph based model of Yan \etal \cite{yan2018spatial} even when using $J_{loc}$ as the feature at each vertex. The graph construction in \cite{yan2018spatial} uses a spatial partitioning scheme for their final model which divides the skeleton graph egde set into several partitions, while the vertex set has no partitions and contains all the joints. The difference in our model is that we divide the \textit{entire} skeleton into \textit{smaller} parts similar to human body parts and hence we use different edge set and vertex set for each part. Compared to graph based model of Li \etal \cite{li2018spatio}, our model performs significantly better on NTURGB+D as well as HDM05. However, it is possible that this is because the number of layers in the network in \cite{li2018spatio} is much smaller (2 vs 9) compared to our model. Our model outperforms both the previous graph based models proposed for skeleton action recognition on the two datasets.

\paragraph{\textsc{Geometric + Kinematic signals}:} Providing an explicit cue to a convolutional network, such as optical flow when performing action recognition from RGB videos \cite{NIPS2014_5353}, which is significant for the task at hand helps learn a richer representation by focusing on the cue. This motivates the use of geometric and kinematic features for skeletal action recognition. For the final configuration of our model, we concatenate the geometric and kinematic signals.

\paragraph{(a) Kinematic: temporal displacements} Temporal displacements provide information about the amount of motion happening between two frames. This information is synonymous to 3D scene flow of a very sparse set of points. We hypothesize that these displacements provide explicit motion information (like optical flow) which makes the model consider displacements as strong features and learn from them. Improvement in performance using this signal can be seen from Table \ref{tab:ablation}(b), for both four-part as well as one-part model across both splits of NTURGB+D.

\paragraph{(b) Geometric: relative coordinates} These provide translation invariant features as explained in \cite{verma2018feastnet} and they have been used effectively to encode skeletons by Ke \etal \cite{ke2017new} into images. Also, Zhang \etal \cite{zhang2017geometric} used relative coordinates as a geometric feature which performs much better than 3D joint locations using a simple stacked LSTM network. We can see improvements in performance provided by relative coordinates in Table \ref{tab:ablation}(b) for both global (one part) and four part-based models, which are the worst and best performing models according to Table \ref{tab:ablation}(a).

\begin{table}[t]
    \begin{center}
        \small
        \begin{tabular}{cc}
            (a) NTURGB+D & (b) HDM05 \\
            \begin{tabular}{ccc}
            \toprule
            \multirow{2}{*}{\textbf{Methods}} & \multicolumn{2}{c}{\textbf{Accuracy}} \\
            & CS & CV \\
            \midrule
            ST Attention \cite{song2017end} & 73.4 & 81.2 \\
            GCA-LSTM \cite{liu2017global} & 74.4 & 82.8 \\
            TCN \cite{kim2017interpretable} & 74.3 & 83.1 \\
            VA-LSTM \cite{zhang2017view} & 79.4 & 87.6 \\
            CNN + MTLN \cite{ke2017new} & 79.6 & 84.8 \\
            \midrule
            Deep STGC \cite{li2018spatio} & 74.9 & 86.3 \\
            STGCN \cite{yan2018spatial} & 81.5 & 88.3 \\
            \midrule
            PB-GCN & \textbf{87.5} & \textbf{93.2} \\
            \bottomrule
            \end{tabular}
            &
            \begin{tabular}{cc}
                \toprule
                \textbf{Methods} & \textbf{Accuracy} \\
                \midrule
                SPDNet \cite{huang2017riemannian} & 61.45 $\pm$ 1.12 \\
                Lie Group \cite{vemulapalli2014human} & 70.26 $\pm$ 2.89 \\
                LieNet \cite{huang2017deep} & 75.78 $\pm$ 2.26 \\
                P-LSTM \cite{Shahroudy_2016_CVPR} & 73.42 $\pm$ 2.05 \\
                \midrule
                Deep STGC \cite{li2018spatio} & 85.29 $\pm$ 1.33 \\
                STGCN \cite{yan2018spatial} & 82.13 $\pm$ 2.39 \\
                \midrule
                PB-GCN & \textbf{88.17} $\pm$ \textbf{0.99} \\
                \bottomrule
            \end{tabular}
        \end{tabular}
    \end{center}
    \caption{\small{Performance comparison with previous methods on two benchmark datasets. The top group of results correspond to non-graph based methods and the middle corresponds to GCN based methods. PB-GCN is our part-based graph convolutional network. Evaluation protocols used: CS (Cross Subject) and CV (Cross View) for NTURGB+D \cite{Shahroudy_2016_CVPR}; 10-fold cross sample validation for HDM05 \cite{huang2017riemannian}.}}
    \label{tab:sota}
\end{table}

\subsection{Comparison to state of the art}
\label{sec:5_3}
\paragraph{\textsc{NTURGB+D}:} For this dataset, we outperform all previous state-of-the-art methods by a large margin. Even without using the signals introduced in section \ref{sec:signals}, we outperform the previous methods which can be seen in Table \ref{tab:ablation}(b) ($J_{loc}$ results). We outperform the previous state-of-the-art graph based method of Yan \etal \cite{yan2018spatial} (STGCN) which is also the state-of-the-art for skeleton based action recognition to the best of our knowledge, by a margin of \textasciitilde6\% and \textasciitilde5\% for the two protocols.

\paragraph{\textsc{HDM05}:} This is a \textasciitilde20x smaller dataset compared to NTURGB+D but contains more than twice the number of classes in NTURGB+D. The length of sequences in this dataset is longer and some of the action classes have only one sequence \cite{Cho2014ClassifyingAV}. Using the protocol of \cite{huang2017riemannian} is therefore very challenging, on which we obtain state-of-the-art results using our model. We outperform the previous state-of-the-art Deep STGC \cite{li2018spatio}, which is a network based on spectral graph convolutions for skeleton action recognition by \textasciitilde3\% at the mean accuracy.
\section{Conclusion}
\label{sec:conclusion}
In this paper, we define a partition of skeleton graph on which spatio-temporal convolutions are formalized through a part-based GCN for the task of action recognition. Such a part-based GCN learns the relations between parts and understands the importance of each part in human actions more effectively than a model that considers entire body as a single graph. We also demonstrate the benefit of giving explicit cues to the convolutional model which are significant from the point of view of the task at hand, such as relative coordinates and temporal displacements for skeletal action recognition. As a result, our model achieves state-of-the-art performance on two challenging action recognition datasets. As a future work, we would like to explore the use of part-based graph model for tasks other than action recognition, such as object detection, measuring image similarity, etc.

\bibliography{egbib}

\end{document}


\maketitle

In this document, we present findings from further quantitative analysis on the action recognition results. Specifically, we compute the confusion matrices of the performance of different models and explain the useful model properties based on our observations. We find that graph-based models can understand actions which involve more motion better than those where skeleton motion is very less and contains object interactions. We also show the importance of using geometric and kinematic features instead of 3D joint locations by performing an experiment on graph-based model of Yan \etal \cite{yan2018spatial}.

\begin{figure}
\begin{center}
    \includegraphics{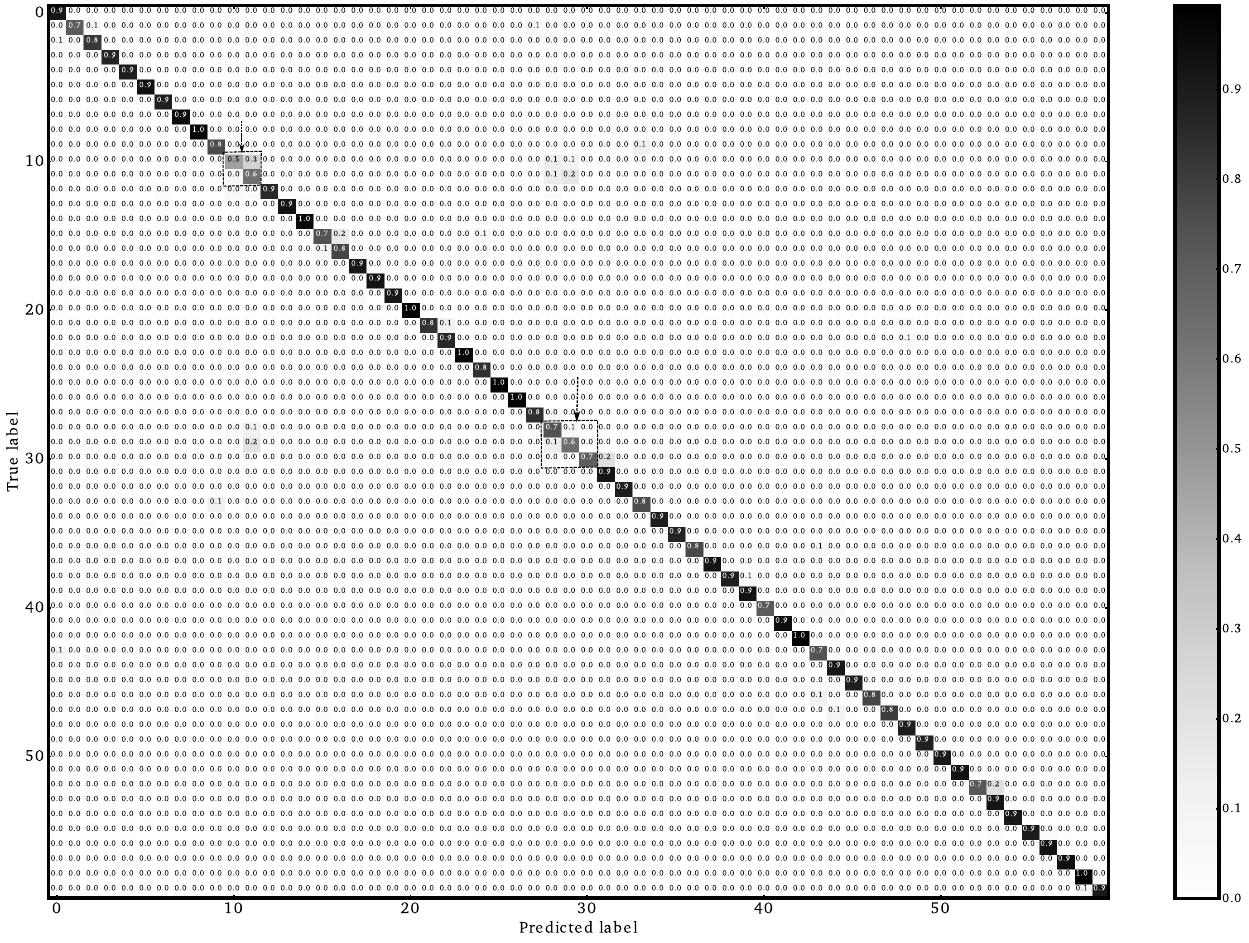}    
\end{center}
\caption{Confusion matrix for model with one part and combined geometric + kinematic features as input.}
\label{fig:onepart}
\end{figure}

\begin{figure}
\begin{center}
    \includegraphics{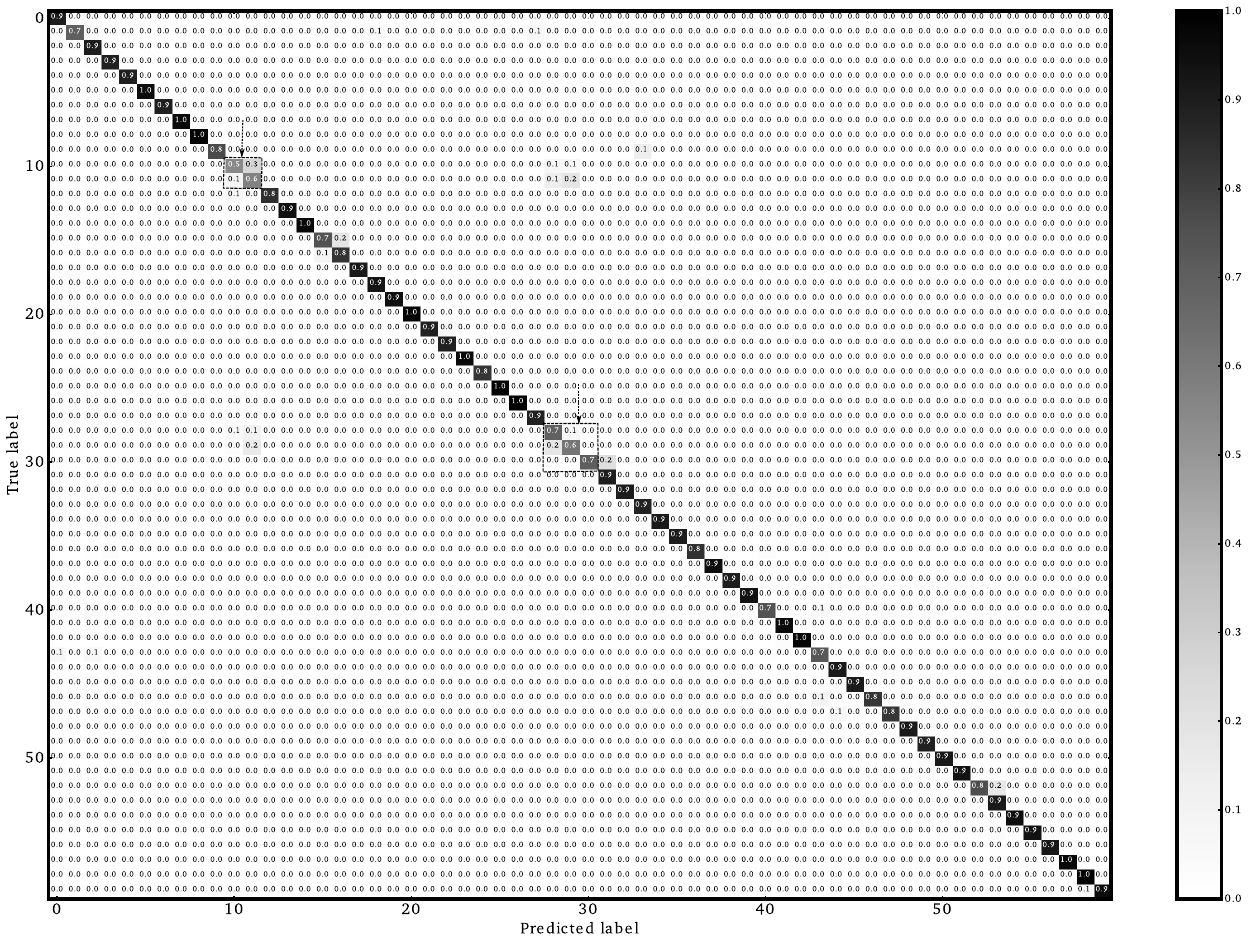}
\end{center}
\caption{Confusion matrix for model with four parts and combined geometric + kinematic features as input.}
\label{fig:fourpart}
\end{figure}

\begin{figure}
\begin{center}
    \includegraphics{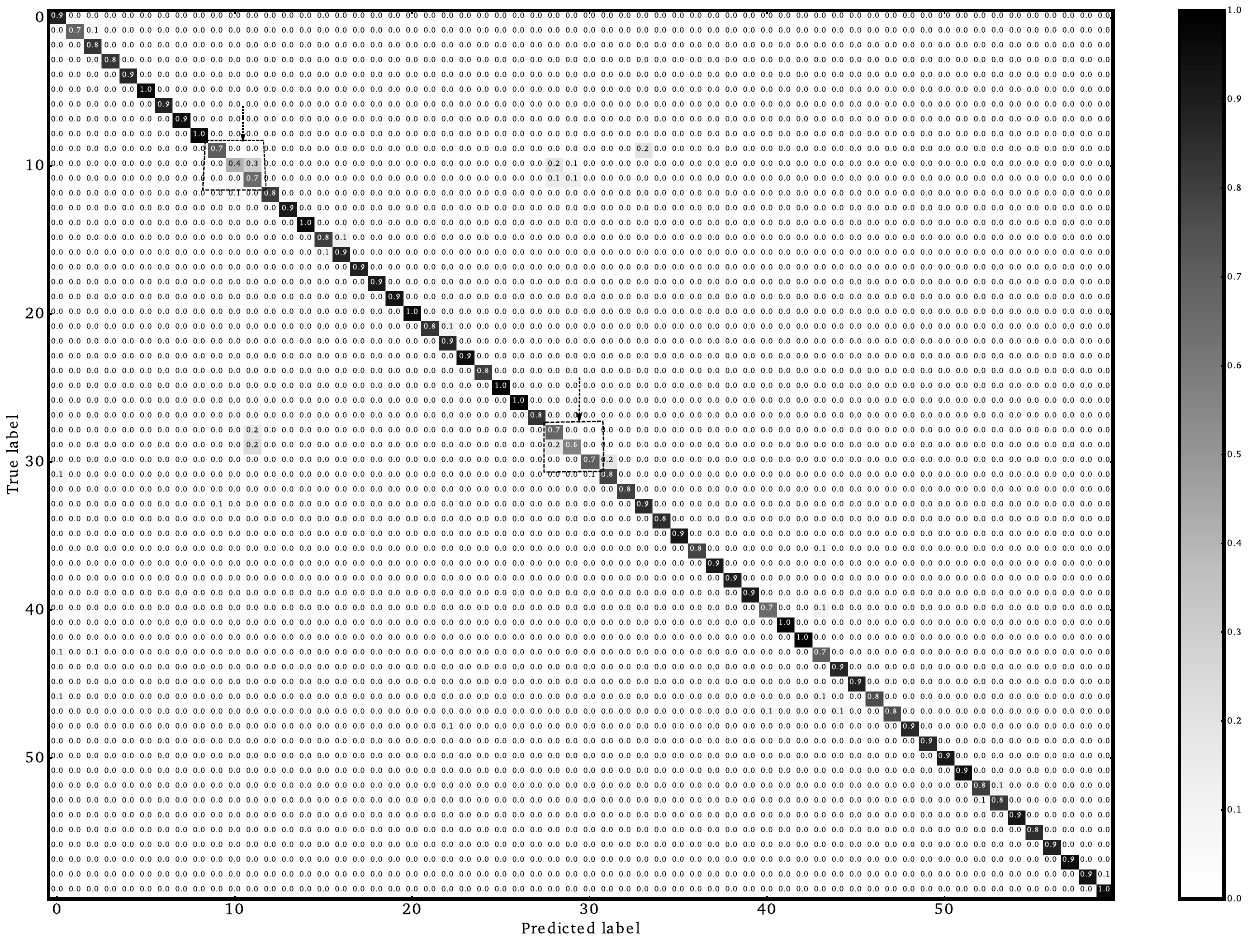}
\end{center}
\caption{Confusion matrix for Yan's graph-based model \cite{yan2018spatial} having 3D joint locations as input signals.}
\label{fig:yan}
\end{figure}

\section{Quantitative Analysis}
We compute the confusion matrices for performance of our part-based graph model, graph model using only one part and Yan's graph model \cite{yan2018spatial}. We did not include Li's graph model \cite{li2018spatio} as no code has been provided by the authors to reproduce the results. The performance for cross subject (CS) evaluation protocol is considered as it is more challenging than the cross view (CV) evaluation protocol. The confusion matrices for different models are shown in Figure \ref{fig:onepart} (model-1), \ref{fig:fourpart} (model-2) and \ref{fig:yan} (model-3). The recognition accuracy for each of these models for cross subject (CS) evaluations is $85.6$, $87.5$ and $81.5$ respectively. The model corresponding to Figure \ref{fig:onepart} is a one-part graph model which does not divide the skeleton graph into parts and it takes a combination of relative joint coordinates $\mathbf{D}_{R}$ and temporal displacements $\mathbf{D}_{T}$ as input. The model corresponding to \ref{fig:fourpart} is our four-part graph model with $\mathbf{D}_{R}$ and $\mathbf{D}_{T}$ as input. Finally, Figure \ref{fig:yan} corresponds to graph-based model introduced in Yan \etal \cite{yan2018spatial} for skeleton action recognition. We proceed to identifying the action classes for which the recognition performance is bad, explain what the reasons are for such performance, propose a possible solution and then compare performance across different classes for models with respect to model-2. 

\begin{table}[t]
  \begin{center}
      \begin{tabular}{|c|cc|}
      \hline
        \multirow{2}{*}{\textbf{Model}} & \multicolumn{2}{|c|}{\textbf{Accuracy}} \\
        \cline{2-3}
        & CS & CV \\
        \hline
        Yan \cite{yan2018spatial} (model-2) & 81.5 & 88.3 \\
        \hline
        Model-2 + $\mathbf{D}_{R} || \mathbf{D}_{T}$ & 86.3 & 92.1 \\
        \hline
      \end{tabular}
  \end{center}
  \caption{Results on NTURGB+D for model-2 \cite{yan2018spatial}, with and without the combined signal $\mathbf{D}_{R} || \mathbf{D}_{T}$ (relative coordinates and temporal displacements).}
  \label{tab:yan}
\end{table}

\subsection{Commonly confused classes}\label{sec:1_1}
The confusion matrices have boxes marked around certain values. These boxes represent the confused classes which are consistent across all models. For example, one of the boxes is around action classes 11 \& 12, which correspond to ``reading'' and ``writing'' actions. These actions are mostly confused amongst each other and also with actions such as ``playing with the phone / tablet'' or ``typing on a keyboard'' (actions 29 \& 30 present in the other marked box) which is clear from the confusion matrices. In all these actions, there is almost no skeleton motion and the differences are manifested in the form of interaction with different objects. Due to these properties, models using skeleton information for recognizing actions give lower performance for these action classes as they do not have access to object information. A possible approach to overcome this limitation on recognition potential is to use RGB information along with skeleton information in order to get information about objects as well.

\subsection{Model-1 vs Model-2}\label{sec:1_2}
Model-2 improves over Model-1 by using a part-based graph representation instead of considering the entire graph as one part. Model-2 achieves better recognition performance by improving over action classes such as ``brushing teeth'' (class 3), ``cheer up'' (class 22), ``make a phone call/answer phone'' (class 28), etc. These actions have a strong correlation with movement of both hands and legs. Due to this correlation, our part-based graph model is able to achieve better performance as it learns from these parts specifically and uses an intuitive way to divide the human body into parts. Being agnostic to parts in human skeleton helps in learning a global representation but learning importance of parts using such a model is difficult, compared to a part-based model.

\subsection{Model-3 vs Model-2}\label{sec:1_3}
Spatio-temporal model of Yan \etal \cite{yan2018spatial} confuses the action of ``clapping'' as well along with the actions mentioned in section \ref{sec:1_1}. The model proposed by Yan \cite{yan2018spatial} partitions the edge set and uses the same vertex set for each partition of edge set. We believe that their model learns the importance of different edges in the skeleton graph and does not learn the importance of parts like our part-based graph model. In order to understand the influence of geometric and kinematic signals as input to a graph-based model, we use the signals on top of model-3 and we find that we get a boost in recognition performance for model-3. The recognition accuracy on NTURGB+D is shown in Table \ref{tab:yan}. This experiment shows that the signals help in improving recognition performance for different graph-models for skeleton action recognition.

\section{Conclusion}
Using a part-based model works better than using a model that does not partition the skeleton graph. However, using only skeletal data for action recognition is not enough as different actions might have similar dynamics of parts in the skeleton but different object interactions. In such cases, RGB information can be used to disambiguate interactions with objects. Providing the network with a cue that is known apriori to work well for the task at hand, viz. relative coordinates and temporal displacements for skeletal action recognition, can improve recognition performance by a large amount as we show in our experiment on previous state-of-the-art model for NTURGB+D \cite{yan2018spatial}.

\bibliography{suppbib}